\documentclass[lettersize,journal]{IEEEtran}
\usepackage{amsmath,amsfonts}
\usepackage{algorithmic}
\usepackage{algorithm}
\usepackage{array}
\usepackage{caption}
\usepackage{subcaption}
\usepackage{textcomp}
\usepackage{stfloats}
\usepackage{url}
\usepackage{verbatim}
\usepackage{graphicx}
\usepackage{cite}
\usepackage{multicol}
\usepackage{color}
\begin{document}

\title{Cluster-Aware Multi-Round Update for Wireless Federated Learning in Heterogeneous Environments}

\author{Pengcheng Sun,
Erwu Liu, \IEEEmembership{Senior Member,~IEEE,}
Wei Ni, \IEEEmembership{Fellow,~IEEE,}
Yuanzhe Geng,
Kanglei Yu,
Rui Wang, \IEEEmembership{Senior Member,~IEEE,}
and Abbas Jamalipour, \IEEEmembership{Fellow,~IEEE}
\thanks{P. Sun, E. Liu, Y. Geng, K. Yu, and R. Wang are with the College of Electronics and Information Engineering, Tongji University, Shanghai 201804, China, E-mails: pc\_sun2020@tongji.edu.cn, erwu.liu@ieee.org, yuanzhegeng@tongji.edu.cn, 2152206@tongji.edu.cn, ruiwang@tongji.edu.cn.
}
\thanks{W. Ni is with the Commonwealth Science and Industrial Research Organization (CSIRO), Marsfield, NSW 2122, Australia, E-mail: wei.ni@data61.csiro.au.
}
\thanks{A. Jamalipour is with the School of Electrical and Computer Engineering, The University of Sydney, Sydney, NSW 2006, Australia, E-mail: a.jamalipour@ieee.org.
}
\thanks{This work is supported in part by grants from the National Natural Science Foundation of China (No. 42171404, No.42225401) and Shanghai Engineering Research Center for Blockchain Applications And Services (No. 19DZ2255100).
}
\thanks{Corresponding author: Erwu Liu.}
}



\maketitle

\begin{abstract}
The aggregation efficiency and accuracy of wireless Federated Learning (FL) are significantly affected by resource constraints, especially in heterogeneous environments where devices exhibit distinct data distributions and communication capabilities. 
This paper proposes a clustering strategy that leverages prior knowledge similarity to group devices with similar data and communication characteristics, mitigating performance degradation from heterogeneity.
On this basis, a novel Cluster-Aware Multi-round Update (CAMU) strategy is proposed, which treats clusters as the basic units and adjusts the local update frequency based on the clustered contribution threshold, effectively reducing update bias and enhancing aggregation accuracy. 
The theoretical convergence of the CAMU strategy is rigorously validated. Meanwhile, based on the convergence upper bound, the local update frequency and transmission power of each cluster are jointly optimized to achieve an optimal balance between computation and communication resources under constrained conditions, significantly improving the convergence efficiency of FL. 
Experimental results demonstrate that the proposed method effectively improves the model performance of FL in heterogeneous environments and achieves a better balance between communication cost and computational load under limited resources.
\end{abstract}


\section{Introduction}
\IEEEPARstart{F}{ederated} Learning (FL) is a promising paradigm that enables efficient distributed machine learning while preserving data privacy \cite{sun2024reconfigurable}. Heterogeneous environments, including Non-Independent and Identically Distributed (Non-IID) data \cite{ma2024fractal, huang2023active}, heterogeneous communication quality \cite{10945769}, and varying computational capabilities \cite{ma2021fedsa, zhang2018adaptive}, severely affect the convergence performance of FL. The efficient deployment of FL in resource-constrained environments remains a significant challenge \cite{ma2021adaptive, wang2017enorm}.

In practice, FL requires devices to frequently transmit models to the parameter server, resulting in significant consumption of communication resources, especially in environments with limited communication and computational capabilities \cite{xu2022adaptive}. The traditional synchronous learning paradigm requires all devices to upload their model parameters to the server simultaneously after completing local training within a global update round, which leads to potential resource waste \cite{zheng2017asynchronous, lian2015asynchronous, agarwal2011distributed}. To address this, researchers propose the Local Multi-round Updates (LMU) strategy, which allows devices to perform multiple rounds of local training before uploading model updates, reducing communication frequency and improving FL efficiency.

Stich et al. \cite{stich2018local} provided a theoretical analysis of the convergence rate with respect to the local update frequency, i.e., the number of local iterations between two consecutive global aggregations. Wang and Luo et al. \cite{wang2019adaptive, luo2021cost} adaptively determined appropriate local update frequencies for participating devices based on resource constraints and dynamic contexts, accelerating training and achieving communication-efficient FL. However, these methods considered only resource constraints and neglected synchronization barriers, leading to non-negligible waiting times and reduced convergence rates. Wang and Yu et al. \cite{wang2019adaptive, yu2019parallel} proposed communication-efficient distributed SGD algorithms with fixed local update frequencies. However, due to the lack of consideration for constrained computational resources and system heterogeneity, these approaches are unsuitable for practical FL environments. Yang et al. \cite{yang2021achieving} pointed out that employing large local update frequencies during training increases the system noise, which may cause the model to converge to a local rather than global optimum \cite{reisizadeh2020fedpaq}. Shi et al. \cite{shi2022toward} investigated how to achieve energy-efficient FL on mobile devices by balancing energy consumption between local updates and model transmission to enhance overall energy efficiency.

FL systems typically involve heterogeneous devices with significant differences in computational capabilities (e.g., CPU frequency, battery life) or communication capabilities (e.g., bandwidth, transmission accuracy). Existing methods usually assign identical or fixed local update frequencies to all devices, which causes faster devices to wait for slower ones during global updates, inevitably lowering global update efficiency. Therefore, assigning personalized local update frequencies to heterogeneous devices is essential to reduce waiting time. Liu et al. \cite{liu2025adaptive} proposed the lightweight FL framework, named Heroes, which combines techniques of enhanced neural composition and adaptive local updates to optimize parameter allocation and update frequency through a greedy algorithm, improving training efficiency in heterogeneous environments. Pan et al. \cite{pan2024time} proposed a time-sensitive FL mechanism that employs deep reinforcement learning to dynamically decide training intensity and uses an optimal deterministic algorithm to adaptively assign local update frequency, improving training efficiency and reducing global aggregation delay. However, these approaches overlook an important issue that heterogeneity in data distribution and communication quality can lead to severe local update bias accumulation when employing LMU strategies in FL systems. Hence, the heterogeneity of devices in terms of data distribution and communication quality must be considered when designing personalized local update strategies.

Clustering devices based on their heterogeneous characteristics and using clusters as the basic update units is an effective way to mitigate device heterogeneity. Zhou et al. \cite{zhou2020petrel} proposed a community-aware synchronous parallel mechanism, performing local updates at the community level to balance convergence efficiency and communication overhead. Zhang et al. \cite{zhang2023fedmds} proposed a clustering strategy based on update delay and direction to schedule devices. Devices in each cluster perform asynchronous updates, followed by synchronous aggregation once the model divergence-based trigger condition is met, addressing the straggler problem and improving aggregation efficiency. Lee et al. \cite{lee2022data} clustered devices by data distribution into nearly IID groups and used multi-armed bandit techniques to select the cluster with the shortest convergence time for training, mitigating the impact of non-IID data, optimizing device selection, and improving aggregation performance. Zheng et al. \cite{zheng2024aou} clustered users based on importance, grouping those with similar data characteristics and update processes to perform asynchronous local updates within clusters and synchronous aggregation across clusters at the base station (BS), which reduces communication time and improves convergence and aggregation efficiency. Liu et al. \cite{liu2024adaptive} proposed an adaptive user clustering strategy based on computational capacity and wireless resources. Devices in the same cluster perform synchronized local updates, while the server executes asynchronous global aggregation, reducing the impact of delayed devices on convergence and improving training efficiency and aggregation performance. These studies offer valuable insights for designing cluster-based customized local update strategies to improve aggregation efficiency in heterogeneous FL.

However, while cluster-based LMU improves aggregation efficiency, it still lacks in ensuring aggregation quality. On the one hand, the local update bias between clusters caused by data heterogeneity still exists. If the clusters with small contribution have multiple rounds of local updates, the global update bias will increase. Therefore, it is important to quantify the heterogeneous degree of cluster contribution. On the other hand, transmission accuracy affects aggregation quality, so the transmission reliability of inter-cluster aggregation must be ensured. Moreover, communication and computation resources should be further balanced to achieve optimal resource allocation under total energy constraints.

Based on the above considerations, we first propose a clustering strategy that groups devices with similar data distributions and communication capabilities into the same cluster, treating each cluster as a new aggregation entity. Then, we design a Cluster-Aware Multi-round Update (CAMU) strategy. By quantifying the contribution of clusters and setting a threshold for it, only those clusters whose contribution exceeds the threshold are allowed to perform multiple rounds of local updates in each round of global aggregation. While other clusters perform only one local update. Finally, we rigorously analyze the convergence of this strategy and formulate a joint optimization problem of cluster local update frequency and transmission power based on the convergence upper bound. Under a total energy constraint, we optimize the allocation of computation and communication resources to tighten the convergence upper bound and enhance FL aggregation performance. Specifically, the contributions of this work are as follows:

\begin{itemize}
    \item A new device clustering mechanism is proposed based on data and communication heterogeneity, which effectively improves the aggregation efficiency of wireless FL.
\end{itemize}

\begin{itemize}
    \item A novel CAMU strategy is designed with clusters as the basic update units, and a cluster contribution threshold is used to avoid the adverse effect of local update bias accumulation on global update.
\end{itemize}

\begin{itemize}
    \item The convergence upper bound of the proposed CAMU strategy is analyzed, and a joint optimization problem of cluster local update frequency and transmission power is formulated under energy consumption constraints, achieving an optimal balance between computation and communication resources under the resource-constrained conditions.
\end{itemize}

We experimentally assess the effectiveness of the clustering strategy in improving FL performance under various Non-IID conditions. While its performance gains are less pronounced than those of our previously proposed DSC strategy\cite{10945769}, it reduces the heterogeneity of update units and lays the foundation for the CAMU strategy. Furthermore, we compare the proposed CAMU strategy with both the traditional local multi-round update method and state-of-the-art optimized variants under resource-sufficient and resource-constrained conditions. The results demonstrate the superiority of CAMU and highlight the necessity of jointly optimizing communication and computation resources in constrained environments.

The remainder of this paper is structured in the following manner: Section II describes the system model, including the learning and wireless communication networks. Section III describes the proposed CAMU strategy and analyzes the convergence of FL under this update mechanism. Section IV presents a joint optimization problem based on the FL convergence analysis under resource constrained conditions, and delineates the solution.  Section V describes the simulations.  Section VI provides the conclusions. 

\textit{Notation}: Upper- and lower-case boldface letters denote matrices and vectors, respectively; $\mathbb{R}^n$ denotes the $n$-dimensional real vector space; $\mathbb{C}^{n_1 \times n_2}$ denotes the $n_1 \times n_2$-dimensional complex space;  $\left|\cdot \right|$ denotes modulus; $\|\cdot\|$ denotes Euclidean norm; $\nabla(\cdot)$ and $\nabla^2(\cdot)$ take gradient and Laplacian, respectively; $diag\left( \cdot \right)$ stands for a diagonal matrix; $\mathbb{E}\left(\cdot\right)$ takes mathematical expectation; $\left(\cdot\right)^\top$ and $\left(\cdot\right)^H$ stand for transpose and conjugate transpose, respectively.

\section{System Model}

We consider an FL system consisting of a BS equipped with $N_a$ antennas, serving as the parameter server, and $K$ single-antenna devices. The $k$-th device ($k = 1, 2, \ldots, K$) has its own local data set $\mathcal{D}_k$. Consider an FL algorithm with the input data vector $\boldsymbol{x}_{ks} \in \mathbb{R}^d$ and the corresponding output $y_{ks} \in \mathbb{R}$, where $s\in \{1,\cdots,|\mathcal{D}_k|\}$ is the index of a data sample. Let $\boldsymbol{w}_k$ denote the local model parameters trained on the $k$-th device.

\subsection{Learning Model}

To minimize the global loss function, FL performs multiple rounds of gradient aggregation until convergence. In the $t$-th aggregation round, the local gradient of the model $\boldsymbol{w} \in \mathbb{R}^q$ (where $q$ is the size of the model parameters) on dataset $\mathcal{D}_k$ is given by:
\begin{equation}
\nabla F_k \left( \boldsymbol{w}^{[t]} \right) = \frac{1}{|\mathcal{D}_k|} \sum_{(\boldsymbol{x}_{ks}, y_{ks}) \in \mathcal{D}_k} \nabla f_k \left(\boldsymbol{x}_{ks}, y_{ks}; \boldsymbol{w}^{[t]} \right),
\end{equation}
where $\nabla f_k \left(\boldsymbol{x}_{ks}, y_{ks}; \boldsymbol{w} \right)$ denotes the gradient of the loss function for the $s$-th sample.

In this paper, the devices in the FL system are divided into $C$ clusters. During each global aggregation round, devices within each cluster first aggregate their gradients at the cluster leader, and then the BS aggregates the gradients from group leaders. The aggregation is computed as follows:
\begin{equation}
\nabla F\left( \boldsymbol{w}^{[t]} \right) = \sum_{c=1}^{C} G_c \left[ \sum_{k=1}^{K_c} G_k \nabla F_k \left( \boldsymbol{w}^{[t]} \right) \right],
\end{equation}
where $c = 1, 2, \ldots, C$ denotes the cluster index, $G_k$ is the intra-group aggregation weight of the $k$-th device, $G_c$ is the inter-cluster aggregation weight of the $c$-th cluster leader, and $K_c$ is the number of devices in the $c$-th cluster, satisfying $K = \sum_{c=1}^{C} K_c$.

The aggregation strategy proposed in this paper requires that devices with similar data distributions be clustered into the same group, so that intra-cluster aggregation approximates FL under IID condition, which is different from that in \cite{10945769}. Accordingly, devices within a group aggregate their local model parameters at the group leader based on the aggregation weight $G_k = \frac{|\mathcal{D}_k|}{\sum_{k=1}^{K_c} |\mathcal{D}_k|}$.

As the data distributions across clusters still exhibit significant heterogeneity, we construct aggregation weights based on the Wasserstein distance \cite{panaretos2019statistical} to address the Non-IID setting. Each cluster is treated as a new entity, where the data distribution of the new entity is defined as the aggregate of the data distributions of devices within the cluster. The Wasserstein distance between two distributions can be computed using the probability mass functions (PMFs) ${pmf}_c$ and ${pmf}_G$, defined as:
\begin{equation}
W_c({pmf}_c, {pmf}_G) = \inf_{\Gamma \in \Pi({pmf}_c, {pmf}_G)} \mathbb{E}_{(x, y) \sim \Gamma} \left[ \|x - y\| \right],
\end{equation}
where ${pmf}_c$ denotes the PMF of the label distribution aggregated from the local datasets of devices in cluster $c$, ${pmf}_G$ denotes the PMF of the global dataset, and $\Gamma$ represents the set of all possible joint distributions. This formulation allows us to quantify the degree of data heterogeneity for each cluster treated as a new entity.

Based on the quantification of data heterogeneity across clusters, we further compute the aggregation weight of cluster $c$ using the Softmax function:

\begin{equation}
G_c = \frac{|\mathcal{D}_c| e^{1 / W_c}}{\sum_{c=1}^{C} |\mathcal{D}_c| e^{1 / W_c}},
\label{Gc}
\end{equation}
where $|\mathcal{D}_c| = \sum_{k=1}^{K_c} |\mathcal{D}_k|$. This defines a stable and smooth aggregation weight. At the same time, this weighting scheme allows relatively homogeneous clusters to contribute more during global updates, while relatively heterogeneous clusters contribute less. This helps balance the influence of each cluster on the global model in Non-IID scenarios, thereby improving the convergence and overall performance of FL.

Finally, the global model update at the BS is given by:
\begin{equation}
\boldsymbol{w}^{[t+1]} = \boldsymbol{w}^{[t]} - \lambda \cdot \nabla F\left( \boldsymbol{w}^{[t]} \right),
\end{equation}
where $\lambda$ is the learning rate.

\subsection{Communication Model}

Let $\boldsymbol{h}_k \in \mathbb{C}^{N_a \times 1}$ be the vector of direct channel coefficients from the $k$-th device to the BS, $h_{jk} \in \mathbb{C}$ be the channel coefficients from the $k$-th device to the $j$-th device. During the model aggregation process of the $t$-th round of communication, the signal received by the BS is as follows.

Let $\boldsymbol{h}_k \in \mathbb{C}^{N_a \times 1}$ be the direct channel coefficient vector from the $k$-th device to the BS, and $h_{jk} \in \mathbb{C}$ be the channel coefficient from the $k$-th device to the $j$-th device. During the $t$-th round of model aggregation, the signal received at the BS is:
\begin{equation}
	\boldsymbol{y}^{[t]} = \sum_{c=1}^{C} \boldsymbol{h}_c p_c \boldsymbol{s}_c^{[t]} + \boldsymbol{n}_0,
\end{equation}
where $p_c \in \mathbb{C}$ is the transmit power scalar of the leader of cluster $c$, $\boldsymbol{s}_c \in \mathbb{C}^{1 \times q}$ is the model parameter aggregated from the cluster members to the leader, and $\boldsymbol{n}_0 \in \mathbb{C}^{N_a \times q}$ is the additive white gaussian noise (AWGN), whose elements follow the distribution $\mathcal{CN}(0, \sigma_{\boldsymbol{n}_0}^2)$, with $\sigma_{\boldsymbol{n}_0}^2$ denoting the environmental noise power.

Assuming all devices follow the CSMA-CA protocol, such as the RTS/CTS mechanism in the IEEE 802.11 standard, concurrent transmissions from multiple devices within the same transmission range are avoided. Communication quality is measured by the signal-to-noise ratio (SNR), which may vary significantly due to the geographical distribution of devices. The received SNR from the $k$-th device to the BS is given by $\gamma_k = \frac{p_k |\boldsymbol{h}_k|^2}{\sigma_{n_0}^2}$. Similarly, when the $k$-th device transmits model parameters to the $j$-th device, the received SNR is given by $\gamma_{jk} = \frac{p_k |h_{jk}|^2}{\sigma_{n_0}^2}$.

\section{Cluster-aware multi-round update strategy}

In this section, devices are clustered based on the similarity of communication quality and data distribution firstly, ensuring that devices within each cluster have stable communication conditions and similar data distributions. Then, taking the cluster as the basic update unit, a cluster-aware local multi-round update strategy is proposed, where devices within each cluster perform multiple rounds of local updates before the cluster leader transmits the updated clustered-model to the BS. This hierarchical structure enhances communication stability and mitigates the accumulation of aggregation bias in heterogeneous scenarios. Finally, the theoretical convergence of the CAMU strategy for FL is analyzed.

\subsection{Device Clustering}

\begin{figure}
\centerline{\includegraphics[width=1\linewidth]{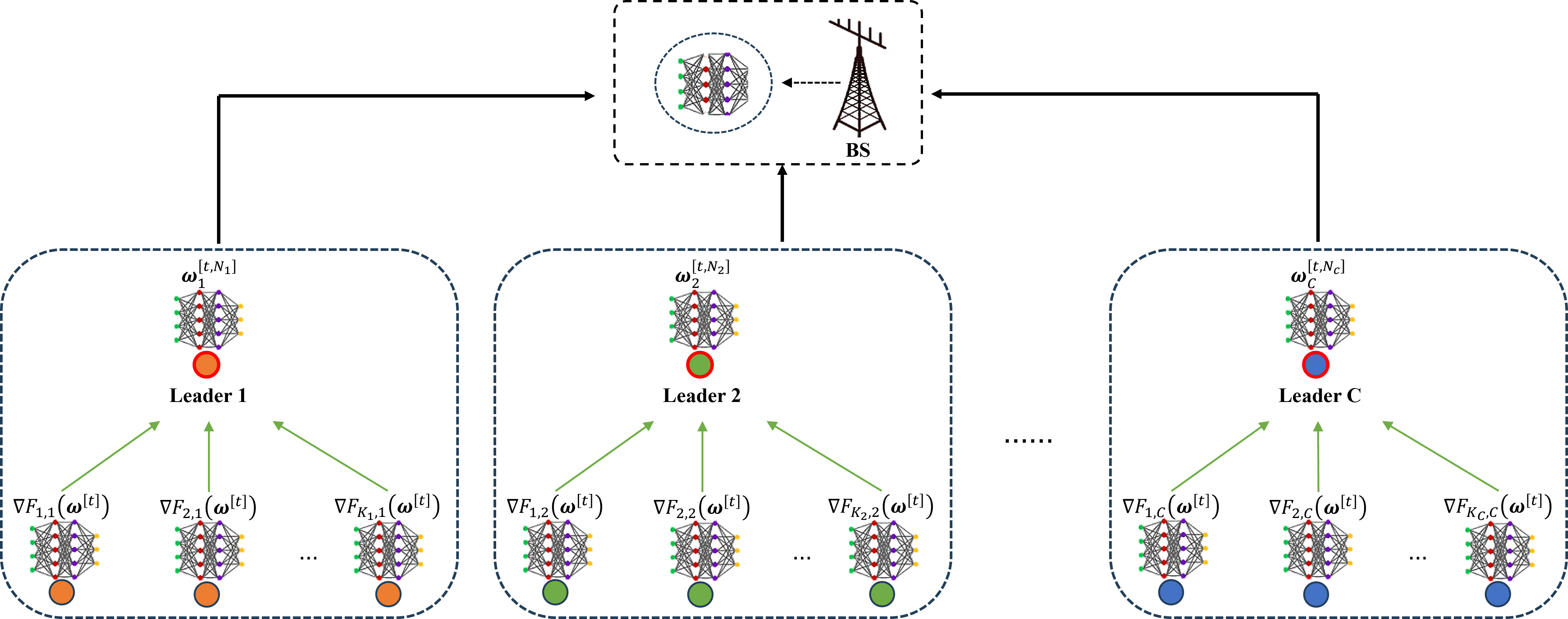}}
\caption{The hierarchical FL architecture based on device clustering, where $K$ devices are divided into $C$ clusters. Local updates are performed by transmitting parameters from cluster members to the leaders, and global updates are performed from the leaders to the BS.}
\label{framework}
\end{figure}

We propose a dual-segment clustering strategy for communication and data heterogeneity. Using the affinity propagation (AP) algorithm, devices are first clustered into primary clusters based on communication quality. An information matrix is then introduced to optimize these primary clusters, ensuring that devices with similar data distributions and communication quality are grouped into the same cluster, which is subsequently treated as a new aggregation entity. This strategy mitigates the impact of communication and data heterogeneity on FL convergence and lays the foundation for the CAMU strategy.
The FL system can thus be updated into a hierarchical structure, as shown in Fig.~\ref{framework}.

To form primary clusters with similar communication quality and ensure accurate transmission of local model parameters, SNR is used as the clustering criterion. It is assumed that device locations and transmission power remain unchanged throughout the training process. The SNR matrix is constructed as:
\begin{equation}
	\boldsymbol{\Gamma} = 
	\begin{bmatrix}
		\gamma_1 & \cdots & \gamma_{1K} \\
		\vdots & \ddots & \vdots \\
		\gamma_{K1} & \cdots & \gamma_K
	\end{bmatrix},
\end{equation}
where $\gamma_i$ denotes the SNR between the $i$-th device and the BS, $\gamma_{ij}$ ($i \neq j$) denotes the SNR between the $i$-th device and the $j$-th device, and $\boldsymbol{\Gamma}$ is a symmetric matrix.

The affinity propagation algorithm\cite{affinity} determines the number of clusters by identifying exemplars—data points that best represent each cluster.  It begins with a similarity matrix reflecting pairwise similarities and iteratively exchanges responsibility and availability messages, where responsibility indicates a point's suitability as an exemplar and availability reflects the appropriateness of selecting it.  This continues until each point is assigned to the exemplar with the highest combined responsibility and availability, forming the clusters.  The algorithm does not require pre-specified cluster numbers and is well-suited for complex, multi-criteria clustering tasks in heterogeneous environments.  We apply the affinity propagation algorithm to the SNR matrix, efficiently constructing primary clusters with similar communication quality and ensuring unambiguous device grouping.

A similarity matrix $\boldsymbol{S}_{c}$ is constructed to describe the similarity between the devices in communication quality, i.e.,
\begin{equation}
    \boldsymbol{S}_{c} = \begin{bmatrix}
        P_{1} & -\gamma^2_{12} & \cdots & -\gamma^2_{1K} \\
        -\gamma^2_{21} & P_{2} & \cdots & -\gamma^2_{2K} \\
        \vdots & \vdots & \ddots & \vdots \\
        -\gamma^2_{K1} & -\gamma^2_{K2} & \cdots & P_{K} 
    \end{bmatrix},
\label{sc}
\end{equation}
where $\{P_1,\cdots,P_K\}$ collects the preference values for communication quality, implying the likelihood of device $k \in \left\{1,\cdots,\ K\right\}$ being the leader of a group and affecting the number of groups. 

A responsibility matrix $\boldsymbol{R}_{c} (i,k)$ is defined to characterize the likelihood of device $k$ serving as the group leader of device $i$. An attribution matrix $\boldsymbol{A}_{c} (i,k)$ is defined to measure the appropriateness of device $i$ nominating device $k$ as its group leader. Both $\boldsymbol{A}_c$ and $\boldsymbol{R}_c$ are initialized as all-zero matrices.

This clustering algorithm iterates over $\boldsymbol{R}_{c} (i,k)$ and $\boldsymbol{A}_{c} (i,k)$ based on the affinity propagation algorithm until the group boundaries do not change for $T_{cl}$ consecutive rounds. Particularly, the responsibility information is updated by 
\begin{equation}
    \boldsymbol{R}_{c}(i,k) = \boldsymbol{S}_{c}(i,k)-\max_{k \neq k'}{[\boldsymbol{S}_{c}(i,k')+\boldsymbol{A}_{c}(i,k')]}.
\label{rc}
\end{equation}
The attribution information is updated by
\begin{equation}
    \boldsymbol{A}_{c}(i,k) \!= \!\min[0,\boldsymbol{R}_{c}(k,k)\!\!+\!\!\!\!\sum_{i' \notin (i,k)}{\max(0,\boldsymbol{R}_{c}(i',k))}],i\!\!\neq \!\! k,
\label{ac1}
\end{equation}
and 
\begin{equation}
    \boldsymbol{A}_{c}(i,i) = \max_{i' \neq k}{[0,\boldsymbol{R}_{c}(i',k)]},i=k.
\label{ac2}
\end{equation}


The responsibility information and attribution information jointly determine the group leaders and members. Specifically, for the $i$-th device, we examine the $i$-th row of the combined matrix $\boldsymbol{C}_{c}=\boldsymbol{R}_{c}(i,k)+\boldsymbol{A}_{c}(i,k)$.  If the maximum of this row is located on the diagonal, the $i$-th device is designated as the group leader corresponding to the column index. If the maximum is not on the diagonal, the $i$-th device is classified as a group member, with the corresponding group leader identified by the column index of the maximum element.

After clustering based on the communication quality, the communication conditions are reasonably consistent within the group. Next, the devices are further clustered according to data heterogeneity within each primary group, so that the devices can contain data with similar label distribution as much as possible in each secondary cluster. 

Suppose that $K$ devices in the FL system possess a total of $\mathcal{D}$ data samples and $\mathcal{L}$ labels. The number of data samples with the $\iota$-th label of the $k$-th device is $\mathcal{C}_k^\iota$. The total number of samples with the $\iota$-th label is $\mathcal{C}^\iota$. The probability that a sample belongs to the $\iota$-th class label in the dataset of the $k$-th device is $P_1 =\mathcal{C}_k^\iota/\mathcal{D}$. The probability of its belonging to the $k$-th device is $P_2 =\mathcal{D}_k/\mathcal{D}$. The probability of its belonging to the $\iota$-th class label is $P_3 =\mathcal{C}^\iota/\mathcal{D}$. Then, a matrix measuring the distribution of the dataset can be written as
\begin{equation}
    \boldsymbol{\Xi} = \begin{bmatrix}
        \frac{\mathcal{C}_1^1}{\mathcal{D}}\log{\frac{\mathcal{D}\mathcal{C}_1^1}{\mathcal{D}_1\mathcal{C}^1}} & \frac{\mathcal{C}_1^2}{\mathcal{D}}\log{\frac{\mathcal{D}\mathcal{C}_1^2}{\mathcal{D}_1\mathcal{C}^2}} & \cdots & \frac{\mathcal{C}_1^\mathcal{L}}{\mathcal{D}}\log{\frac{\mathcal{D}\mathcal{C}_1^\mathcal{L}}{\mathcal{D}_1\mathcal{C}^\mathcal{L}}} \\
        \frac{\mathcal{C}_2^1}{\mathcal{D}}\log{\frac{\mathcal{D}\mathcal{C}_2^1}{\mathcal{D}_2\mathcal{C}^1}} & \frac{\mathcal{C}_2^2}{\mathcal{D}}\log{\frac{\mathcal{D}\mathcal{C}_2^2}{\mathcal{D}_2\mathcal{C}^2}} & \cdots & \frac{\mathcal{C}_2^\mathcal{L}}{\mathcal{D}}\log{\frac{\mathcal{D}\mathcal{C}_2^\mathcal{L}}{\mathcal{D}_2\mathcal{C}^\mathcal{L}}} \\
        \vdots & \vdots & \ddots & \vdots \\
        \frac{\mathcal{C}_K^1}{\mathcal{D}}\log{\frac{\mathcal{D}\mathcal{C}_K^1}{\mathcal{D}_K\mathcal{C}^1}} & \frac{\mathcal{C}_K^2}{\mathcal{D}}\log{\frac{\mathcal{D}\mathcal{C}_K^2}{\mathcal{D}_K\mathcal{C}^2}} & \cdots & \frac{\mathcal{C}_K^\mathcal{L}}{\mathcal{D}}\log{\frac{\mathcal{D}\mathcal{C}_K^\mathcal{L}}{\mathcal{D}_K\mathcal{C}^\mathcal{L}}} 
    \end{bmatrix},
\label{datama}
\end{equation}
where $\frac{\mathcal{C}_k^\iota}{\mathcal{D}}\log{\frac{\mathcal{D}\mathcal{C}_k^\iota}{\mathcal{D}_k\mathcal{C}^\iota}}$ quantifies the information that a sample is classified into the $\iota$-th label of the $k$-th device. Based on $\boldsymbol{\Xi}$, we construct a similarity matrix to describe the distribution of the dataset, as given by
\begin{equation}
    \boldsymbol{S}_{d} = \begin{bmatrix}
        P_{d} & s_{1,2} & \cdots & s_{1,K_{c_{com}}} \\
        s_{2,1} & P_{d} & \cdots & s_{2,K_{c_{com}}} \\
        \vdots & \vdots & \ddots & \vdots \\
        s_{K_{c_{com}},1} & s_{K_{c_{com}},2} & \cdots & P_{d} 
    \end{bmatrix},
\label{sd}
\end{equation}
where $s_{i,k}=\{\sum_{\iota=1}^{\mathcal{L}}{[\boldsymbol{\Xi} (i,\iota)-\boldsymbol{\Xi} (k,\iota)]^2}\}^2$ with $i \neq k$, ${P}_{d}$ is the preference value for the data distribution, and $K_{l_{com}}$ denotes the number of devices in the $l_{com}$-th primary group.

Similarly, we define a responsibility matrix $\boldsymbol{R}_{d} (i,k)$ and an attribution matrix $\boldsymbol{A}_{d} (i,k)$ for the similarity matrix $\boldsymbol{S}_d$ to iteratively update the secondary groups based on the data heterogeneity. The group leaders and members are selected in the same way as in the primary groups. For the remaining ungrouped devices, the Euclidean distance-based proximity principle can be adopted to group them into their respective nearby groups. 

The clustering algorithm in this section groups $K$ devices into $C$ clusters, ensuring that devices within each cluster have similar communication quality and data distributions. In the subsequent aggregation strategy, the cluster serves as the basic update unit for model aggregation. Compared to the traditional device-level update approach, this method reduces communication resource consumption among intra-cluster devices, allowing more resources to be allocated to computation. As a result, it improves global update efficiency and reduces overall energy consumption. This cluster-level update approach maintains model convergence while achieving a better balance between computation and communication resources, enhancing the applicability of FL in resource-constrained environments.

\subsection{CAMU Strategy}

As mentioned before, devices perform the same number of local iterations in each global aggregation round in most existing FL mechanisms. However, due to heterogeneity in computing capabilities, devices with weak computational power often slow down the aggregation process of the global model. To mitigate this straggler problem, an intuitive approach is to assign different training intensities (i.e., local update frequencies) to devices based on their computing capabilities. This paper considers increasing the local update frequency for clusters with strong computing capabilities to fully utilize the idle time of efficient clusters, thereby reducing the global update frequency, lowering system energy consumption, and improving learning efficiency. The computation time of a cluster depends on the device with the longest computation time within that cluster.

\begin{figure}
	\begin{minipage}{0.49\linewidth}
		\centering
		\includegraphics[width=1.0\linewidth]{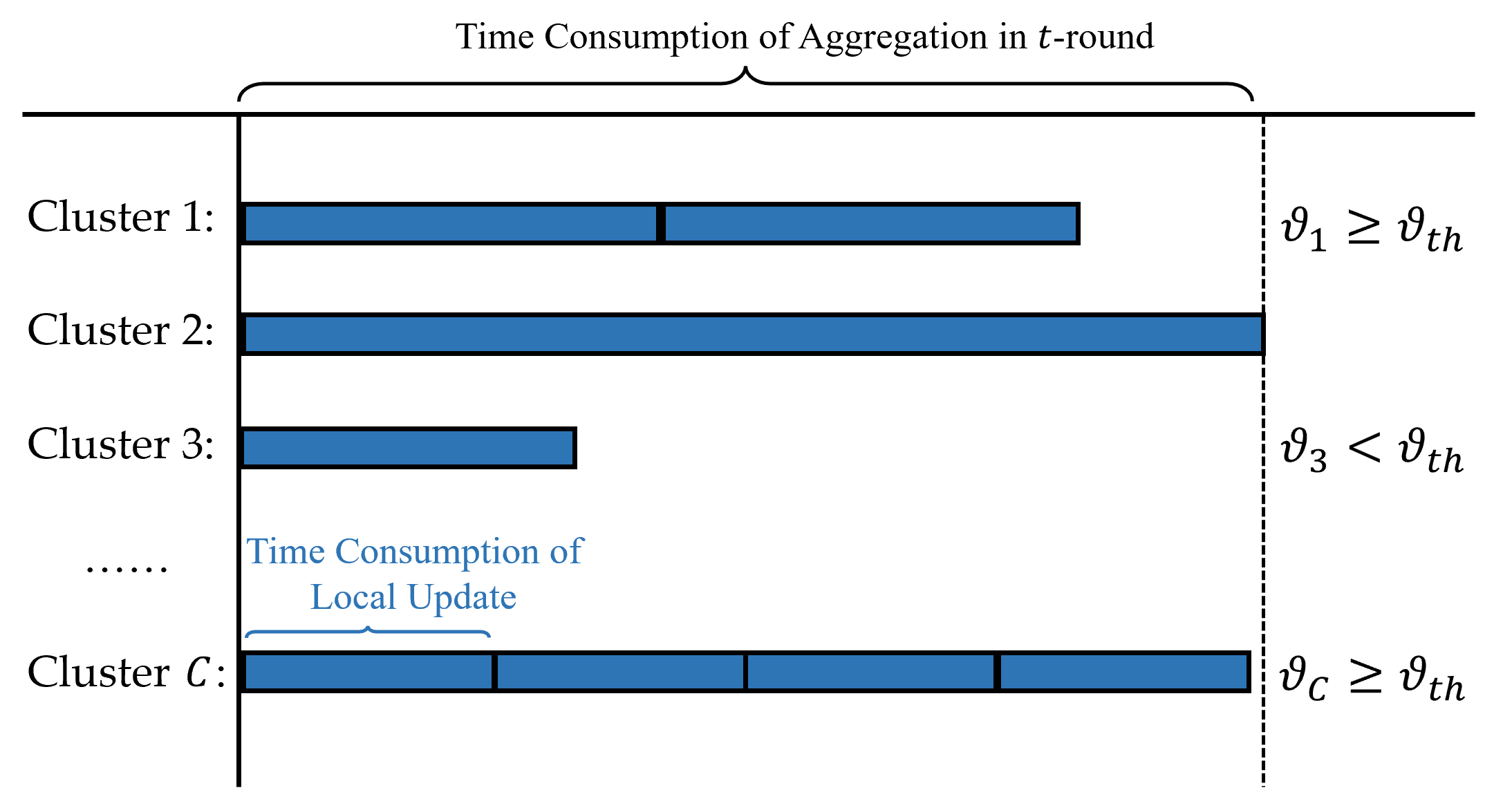}
		\subcaption{}
	\end{minipage}%
	\begin{minipage}{0.4995\linewidth}
		\centering
		\includegraphics[width=1.0\linewidth]{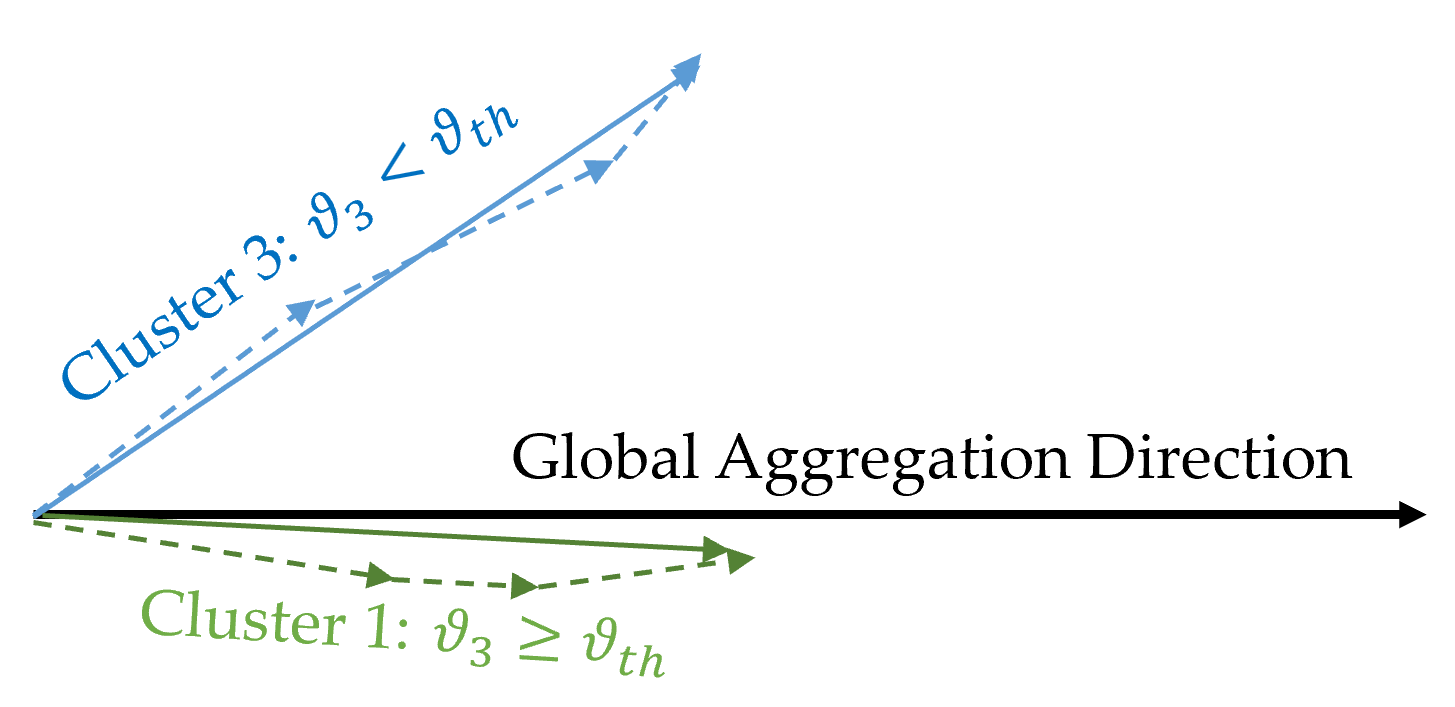}
		\subcaption{}
	\end{minipage}
	\caption{Cluster-based local multi-round update mechanism. (a) illustrates how the mechanism minimizes potential time waste; and (b) shows the impact of clusters with different contribution levels on the global update.}
	\label{LMU}
\end{figure}

However, data heterogeneity poses challenges to the strategy. Although the proposed clustering strategy alleviates the impact of data heterogeneity on aggregation to some extent, it does not fundamentally eliminate the bias accumulation of local update. Differences in data distributions across clusters lead to varying contributions of local updates to the global update, and the bias accumulation of local update mainly arises when clusters with lower contribution are allowed to perform multiple local iterations. Fig.~\ref{LMU} intuitively illustrates the cluster-based local multi-round update mechanism and shows the impact of clusters with different contribution levels on the global update. Clusters with lower contribution levels have greater local update bias; if they are allowed to perform multiple local iterations, their bias can significantly affect the direction of the global update, and degrade aggregation performance.

To reduce the bias accumulation of local update, it is essential to allocate local update frequencies reasonably based on the contribution of each cluster. Thus, we propose the concept of cluster contribution to quantify each cluster’s impact on the global update. This contribution depends solely on the data distribution and is defined as:
\begin{equation}
	\vartheta_c = |\mathcal{D}_c| \cdot e^{1 / W_c},
\end{equation}
where $|\mathcal{D}_c|$ denotes the number of samples in the $c$-th cluster, and $W_c$ represents the Wasserstein distance between the PMF of the cluster’s labels and the PMF of the overall sample labels.

Based on the cluster contribution, the following strategy is proposed: if the contribution $\vartheta_c$ of the $c$-th cluster to the global update is below an acceptable threshold $\vartheta_{th}$, the system allows it to perform only one local update per global aggregation round, regardless of its computing capability; otherwise, it is allowed to perform multiple local updates based on its computing capability. The local update frequency of the $c$-th cluster is defined as:
\begin{equation}
	N_c = 1 + S_c \cdot n_c,
\end{equation}
where
\begin{equation}
	S_c =
	\begin{cases}
		1, & \text{if} \quad \vartheta_c \geq \vartheta_{th} \\[15pt]
		0, & \text{if} \quad \vartheta_c < \vartheta_{th}
	\end{cases}.
\end{equation}

The CAMU strategy mitigates the impact of data heterogeneity on convergence by suppressing excessive training in low-contribution clusters, thus avoiding ineffective updates. The strategy improves the efficiency of global FL updates while effectively alleviating the influence of local update bias, ensuring stable model aggregation under Non-IID conditions. In practical deployments, where computational and communication resources are typically constrained, optimizing resource allocation based on the CAMU strategy is crucial to balance learning performance and energy consumption.

\subsection{Convergence Analysis}

This section provides a theoretical analysis of the convergence performance of FL, focusing on deriving the impact of the proposed hierarchical architecture and the CAMU strategy on global convergence. Through rigorous mathematical derivation, a convergence upper bound related to cluster transmit power and local update frequency is established for hierarchical wireless FL, offering theoretical support for optimizing the allocation of computational and communication resources under constrained conditions to enhance overall system performance.

To analyze the convergence of FL, we introduce assumptions \textbf{A1}–\textbf{A4}:

\textbf{A1} \quad $F_k$ is $L$-smooth, i.e., $\|F_k(\boldsymbol{w}) - F_k(\boldsymbol{w}')\| \leq L \|\boldsymbol{w} - \boldsymbol{w}'\|$. According to this assumption and the triangle inequality, $F(\boldsymbol{w})$ is also $L$-smooth. Additionally, its gradient is Lipschitz continuous;

\textbf{A2} \quad $F_k$ is $\mu$-strongly convex, i.e., $F_k(\boldsymbol{w}^{[t+1]}) \geq F_k(\boldsymbol{w}^{[t]}) + (\boldsymbol{w}^{[t+1]} - \boldsymbol{w}^{[t]})^\top \nabla F(\boldsymbol{w}^{[t]}) + \frac{\mu}{2} \|\boldsymbol{w}^{[t+1]} - \boldsymbol{w}^{[t]}\|^2$;

\textbf{A3} \quad $F(\boldsymbol{w})$ is second-order continuously differentiable. Based on this, together with \textbf{A1} and \textbf{A3}, we have $\mu \boldsymbol{I} \preceq \nabla^2 F(\boldsymbol{w}) \preceq L \boldsymbol{I}$;

\textbf{A4} ($\delta$-local dissimilarity) The local loss function $F_k(\boldsymbol{w}^{[t]})$ at $\boldsymbol{w}^{[t]}$ is $\delta$-locally dissimilar, i.e.,
$\mathbb{E} \left[ \|\nabla F_k(\boldsymbol{w}^{[t]})\|^2 \right] \leq \delta^2 \|\nabla F(\boldsymbol{w}^{[t]})\|^2,  \text{for } k = 1, \ldots, K$,
where $\mathbb{E}$ denotes the weighted aggregation over participating devices, and the larger the $\delta \geq 1$, the more heterogeneous the data, indicating that local updates (i.e., local gradients) are more divergent. When the data distribution is IID, $\delta$ approaches 1. Thus, $\delta$ characterizes the degree of Non-IID-ness.

In the considered FL system, image classification is taken as an example task. Accordingly, the cross-entropy function is chosen as the loss function, which is strongly convex and satisfies the aforementioned assumptions \textbf{A1}–\textbf{A4}. The following theorem provides the convergence upper bound of FL based on the proposed CAMU strategy.

\vspace{2 mm}
\noindent\textbf{Theorem 1:} \textit{Given the optimal global model $\boldsymbol{w}^*$ under ideal channel conditions, inter-cluster local dissimilarity index $\delta$, intra-cluster local dissimilarity index $\delta_c$, intra-cluster aggregation weight $G_k$, inter-cluster aggregation weight $G_c$, learning rate $\lambda$, cluster local update frequency $N_c$, and transmit power allocation $\boldsymbol{p}$, the convergence upper bound of FL is given by:
\begin{equation}
    \begin{aligned}
    	&\mathbb{E} \left[ F(\boldsymbol{w}^{[t+1]}) - F(\boldsymbol{w}^*) \right] \\ &\leq A^T \, \mathbb{E} \left[ F(\boldsymbol{w}^{[0]}) - F(\boldsymbol{w}^*) \right] 
    	+ \frac{1 - A^T}{1 - A} \sum_{c=1}^{C} G_c^2 \cdot \frac{\sigma_n^2}{p_c^2 \| \boldsymbol{h}_c \|^2},
    \end{aligned}
\end{equation}
where $T$ is the total number of aggregation rounds, and $A = 1 + \sum_{c=1}^{C} N_c \left( \mu L \lambda^2 \delta^2 G_c^2 \delta_c^2 \sum_{k=1}^{K_c} G_{k,c}^2 
	- 2\mu \lambda \delta G_c \delta_c \sum_{k=1}^{K_c} G_{k,c} \right)$.
}

\textit{Proof:} See Appendix I.
\vspace{2 mm}

According to \textbf{Theorem 1}, when $0 < A < 1$, the upper bound $\mathbb{E} \left[ F(\boldsymbol{w}^{[t+1]}) - F(\boldsymbol{w}^*) \right]$ converges as rate of $A$. Therefore, $A$ can be regarded as the convergence rate of FL. To ensure the convergence, the following condition should be satisfied:
\begin{equation}
	\lambda < \frac{2 \sum_{k=1}^{K_c} G_{k,c}}{L \delta G_c \delta_c \sum_{k=1}^{K_c} G_{k,c}^2}.
\end{equation}
In other words, when the learning rate $\lambda$ satisfies the inequality, the FL system convergences. It can be seen that the learning rate $\lambda$ is inversely proportional to the degree of data heterogeneity—the more imbalanced the data distribution, the smaller the required $\lambda$ to ensure convergence.

Moreover, since the FL system exhibits a hierarchical architecture in this paper and clusters serve as the basic update units, the local update process can be viewed as one or more approximate IID global updates performed within each cluster. This hierarchical structure requires ensuring not only the convergence of the overall FL system but also the convergence of intra-cluster local updates. Accordingly, we propose the following corollary to clarify the convergence conditions within clusters:

\vspace{2 mm}
\noindent\textbf{Corollary 1:} \textit{According to the aggregation strategy proposed in Section III-A, the devices within a cluster and the cluster leader together form an approximate IID FL subsystem. When the following condition is satisfied:
\begin{equation}
	\delta G_c < 1 < \frac{2 \lambda \delta_c \sum_{k=1}^{K_c} G_{k,c}}{L \lambda^2 \delta_c^2 \sum_{k=1}^{K_c} G_{k,c}^2},
\end{equation}
the intra-cluster local update process is guaranteed to converge.}

\textit{Proof:} See Appendix II.
\vspace{2 mm}

In fact, the convergence condition in \textbf{Corollary 1} is equivalent to the learning rate condition required for the convergence of the overall FL system, indicating that with appropriate learning rate settings, the intra-cluster update process can effectively support the global system's convergence. Moreover, the condition $\delta G_c < 1$ further implies that, during global aggregation, the aggregation weight of each cluster should be inversely proportional to its degree of data heterogeneity, which means that when data heterogeneity is high, the corresponding aggregation weight should be appropriately reduced to mitigate the bias accumulation caused by heterogeneous distributions, thereby improving overall convergence performance.

Furthermore, based on the proposed CAMU strategy, the convergence rate can be divided into two parts, as shown in \eqref{raet A}, where $\vartheta_c < \vartheta_{th}$ corresponds to low-contribution clusters, which have larger heterogeneity factors, i.e., $\delta_{c_{\vartheta_c < \vartheta_{th}}} \geq \delta_{c_{\vartheta_c \geq \vartheta_{th}}}$. The CAMU strategy reduces the local update frequency of such clusters to maximally suppress the accumulation of local bias, thus obtains a smaller $A$ and accelerates the convergence rate.
\begin{figure*}[t]
    \begin{equation}
    \begin{aligned}
    A &= 1 + \sum_{c=1}^{C} N_c \left( \mu L \lambda^2 \delta^2 G_c^2 \delta_c^2 \sum_{k=1}^{K_c} G_{k,c}^2 
		- 2 \mu \lambda \delta G_c \delta_c \sum_{k=1}^{K_c} G_{k,c} \right) \\
		&= 1 +\sum_{c\,:\,\vartheta_c < \vartheta_{th}} \left( \mu L \lambda^2 \delta^2 G_c^2 \delta_{c_{\vartheta_c < \vartheta_{th}}}^2 \sum_{k=1}^{K_c} G_{k,c}^2 
		- 2 \mu \lambda \delta G_c \delta_{c_{\vartheta_c < \vartheta_{th}}} \sum_{k=1}^{K_c} G_{k,c} \right) \\
		&\quad + \sum_{c\,:\,\vartheta_c \geq \vartheta_{th}} (1 + n_c) \left( \mu L \lambda^2 \delta^2 G_c^2 \delta_{c_{\vartheta_c \geq \vartheta_{th}}}^2 \sum_{k=1}^{K_c} G_{k,c}^2 
		- 2 \mu \lambda \delta G_c \delta_{c_{\vartheta_c \geq \vartheta_{th}}} \sum_{k=1}^{K_c} G_{k,c} \right).
    \end{aligned}
    \label{raet A}
    \end{equation}
\hrule
\end{figure*}

The convergence upper bound reveals that higher transmit power $\boldsymbol{p}_c$ or a higher local update frequency $\boldsymbol{n}_c$ both help accelerate convergence and reduce error. It is worth noting that the local update frequency reflects the investment of computational resources, while the transmit power represents the consumption of communication resources. In resource-constrained scenarios, optimizing the allocation of computational and communication resources is beneficial for achieving the optimal balance between performance and resource utilization, further enhancing overall system performance.

\section{Resource optimization based on CAMU}

\subsection{Problem Formulation}

The convergence of the CAMU strategy is jointly influenced by the local update frequency and transmit power. However, in practical systems, computational and communication resources typically face strict budget constraints. Optimizing a single resource may cause system imbalance due to the nonlinear coupling between different resources. Therefore, under total resource constraints, constructing a joint optimization model for computation and communication is a key challenge for achieving efficient FL. This section formulates a joint optimization problem based on the convergence upper bound to coordinate the allocation of computational and communication resources, aiming to maximize the learning efficiency of the global model while ensuring system stability.

According to the convergence upper bound, frequent local updates and transmission errors lead to a performance gap between the current model $F(\boldsymbol{w}^{[t+1]})$ and the optimal model $F(\boldsymbol{w}^*)$, expressed as:
\begin{equation}
	\text{GAP} = \frac{1 - A^T}{1 - A} \sum_{c=1}^{C} G_c^2 \cdot \frac{\sigma_n^2}{p_c^2 \| \boldsymbol{h}_c \|^2}.
\end{equation}

As noted earlier, when $A < 1$, the convergence is guaranteed, and in this case, it is sufficient to minimize the $GAP$. Note that when $T \rightarrow \infty$, the $GAP$ serves as the upper bound of $\mathbb{E} \left[ F(\boldsymbol{w}^{[t+1]}) - F(\boldsymbol{w}^*) \right]$, i.e.,
\begin{equation}
	\lim_{T \rightarrow \infty} \mathbb{E} \left[ F(\boldsymbol{w}^{[t+1]}) - F(\boldsymbol{w}^*) \right] 
	\leq \frac{1}{1 - A} \sum_{c=1}^{C} G_c^2 \cdot \frac{\sigma_n^2}{p_c^2 \| \boldsymbol{h}_c \|^2}.
\end{equation}
The $GAP$ mainly arises from local update bias and transmission errors. If the local update frequency is too low, the communication frequency increases, introducing more transmission errors and biasing resource allocation toward communication overhead. Conversely, if the local update frequency is too high, although the CAMU strategy can alleviate the resulting bias accumulation to some extent, it occupies excessive communication resources, leading to degraded transmission quality and, in turn, exacerbating error accumulation.

In summary, to enhance the convergence performance of FL, it is necessary to balance the allocation of computational and communication resources under a total resource constraint. Therefore, the following optimization problem is formulated:
\begin{subequations}
	\label{problem1}
	\begin{align}
		\mathcal{P}1: \quad & \min_{\boldsymbol{p},\, \boldsymbol{n}} \left[ \frac{1}{1 - A} \sum_{c=1}^{C} G_c^2 \cdot \frac{\sigma_n^2}{p_c^2 \| \boldsymbol{h}_c \|^2} \right], \\
		s.t. \quad & \sum_{c=1}^{C} \left\{ \frac{p_c q}{B_c \log_2(1 + \gamma_c)} + n_c \sum_{k=1}^{K_c} \frac{l_k q}{f_k} p_{cmp} \right\} \leq E_{total},  \\
		& \sum_{k=1}^{K} |p_k|^2 \leq P_{max},
	\end{align}
\end{subequations}
where $B_c$ denotes the transmission bandwidth used by cluster the $c$-th leader, so $\frac{p_c q}{B_c \log_2(1 + \gamma_c)}$ represents the transmission energy consumption; $l_k$ denotes the number of CPU cycles required per training sample for the $k$-th device, and $f_k$ is computing capability, so $n_c \sum_{k=1}^{K_c} \frac{l_k q}{f_k} p_{cmp}$ represents the total computational energy consumption of the $c$-th cluster in one global update. Optimizing the allocation of computational and communication resources helps minimize the $GAP$ and improve convergence performance.

\subsection{Joint Optimization using PPO algorithm}

The problem under the CAMU strategy presents a key challenge: transmit power and local update frequency are strongly coupled continuous variables, making traditional static optimization intractable. Additionally, resource constraints implicitly affect system performance, preventing a closed-form optimal solution. Reinforcement learning (RL), through iterative interaction with a deterministic environment, can effectively learn near-optimal resource allocation strategies, offering superior flexibility under complex constraints.

Since traditional RL methods (e.g., Deep Q-Network, DQN) can not directly handle continuous action spaces,  and discretizing the action space reduces both solution accuracy and efficiency. Therefore, this section adopts the Proximal Policy Optimization (PPO) algorithm to solve the problem. PPO, based on an Actor-Critic framework, generates continuous actions directly and uses a clipped objective function to effectively control policy update steps, achieving stable and efficient optimization.

Specifically, we map the original problem $\mathcal{P}1$ into a reinforcement learning process, with the following key elements:

(1) State. In the $n$-th optimization round, the state vector is defined as:
\begin{equation}
	\boldsymbol{s}_n = \left[ \{p_c^{(n-1)}\}_{c=1}^{C}, \{n_c^{(n-1)}\}_{c=1}^{C}, {GAP}^{(n-1)}, E^{(n-1)} \right],
\end{equation}
where the state vector includes the previous round’s power allocation, clustered local update frequencies, the value of objective function, and actual system energy consumption. This information jointly characterizes the system’s environment prior to the current decision round.

(2) Action. The action in the $n$-th optimization round is defined as the resource allocation for each cluster:
\begin{equation}
	\boldsymbol{a}_n = \left\{ \left( p_c^{(n)}, n_c^{(n)} \right) \right\}_{c=1}^{C}.
\end{equation}
The action values are continuous and are projected during implementation to satisfy resource constraints.

(3) Reward. To align reinforcement learning’s objective of maximizing cumulative reward with the optimization problem’s objective of minimizing the $GAP$, the reward function is defined as the negative of the objective function:
\begin{equation}
	r_n = -{GAP}^{(n-1)}.
\end{equation}
With this design, the agent receives a higher reward when resource allocation reduces the objective function. To ensure compliance with the energy budget $E_{{total}}$ and maximum power constraint $P_{{max}}$, an additional penalty term is introduced when the action results in resource overuse:
\begin{equation}
	r_n = -{GAP}^{(n-1)} - \alpha \left[ E^{(n-1)} - E_{{total}} \right],
\end{equation}
where $\alpha$ is the penalty coefficient. This design helps guide the agent to learn resource configurations that satisfy the constraints.

The training process of the PPO algorithm mainly involves action sampling, advantage calculation, policy update, and value network update. Specifically, given the state $\boldsymbol{s}_n$, the policy network $\pi_\theta$ first generates the probability distribution of the resource allocation (i.e., transmit power and local update frequency) for each cluster in the current round and samples the actual action as $ \boldsymbol{a}_n \sim \pi_\theta(\boldsymbol{a} \mid \boldsymbol{s}_n) $, where $\pi_\theta(\boldsymbol{a} \mid \boldsymbol{s}_n)$ denotes the action probability distribution generated by the policy network conditioned on $\boldsymbol{s}_n$, and $\boldsymbol{a}_n$ is the sampled resource allocation for the current round. Next, the advantage function $A_n$ is computed to evaluate the performance improvement of the selected action $\boldsymbol{a}_n$ over the baseline in terms of the optimization objective:
\begin{equation}
	A_n = r_n + \Upsilon V_\phi(\boldsymbol{s}_{n+1}) - V_\phi(\boldsymbol{s}_n),
\end{equation}
where $r_n$ is the immediate reward of the current optimization round (i.e., the negative convergence error $GAP$ plus any penalty for constraint violation); $V_\phi(\boldsymbol{s})$ is the value estimate of state $\boldsymbol{s}$ from the Critic network; and $\Upsilon \in (0,1)$ is the discount factor balancing current reward and expected future return.

To ensure training stability, PPO introduces the action probability ratio between the new and old policies:
\begin{equation}
	r_n(\theta) = \frac{\pi_\theta(\boldsymbol{a}_n \mid \boldsymbol{s}_n)}{\pi_{\theta_{{old}}}(\boldsymbol{a}_n \mid \boldsymbol{s}_n)}.
\end{equation}
This ratio is used to define the clipped objective function:
\begin{equation}
	L^{{clip}}(\theta) = \mathbb{E}_n \left[ \min \left( r_n(\theta) A_n, \ {clip}(r_n(\theta), 1 - \epsilon, 1 + \epsilon) A_n \right) \right],
\end{equation}
where $\epsilon$ is the clipping threshold (typically set to $0.2$), which limits the step size of policy parameter updates to prevent instability from large policy shifts; the function ${clip}(x, 1 - \epsilon, 1 + \epsilon)$ constrains $x$ within the interval $[1 - \epsilon, 1 + \epsilon]$.

The Critic network updates the state value estimates by minimizing the loss of mean squared error:
\begin{equation}
	L^{{VF}}(\phi) = \mathbb{E}_n \left[ \left( V_\phi(\boldsymbol{s}_n) - (r_n + \Upsilon V_\phi(\boldsymbol{s}_{n+1})) \right)^2 \right],
\end{equation}
where the term $r_n + \Upsilon V_\phi(\boldsymbol{s}_{n+1})$ represents the target value computed based on the immediate reward and the value of the next state.

Finally, the two networks' parameters are alternately updated via gradient descent: the Actor network parameters $\theta$ are updated by maximizing the clipped objective function $L^{{clip}}(\theta)$, while the Critic network parameters $\phi$ are updated by minimizing the loss of value function $L^{{VF}}(\phi)$. These steps are repeated until the predefined maximum number of training rounds is reached. With the clear definitions of state, action, and reward, along with a robust policy update mechanism, the PPO algorithm in this section autonomously learns the optimal combination of transmit power and local update frequency that satisfies resource constraints, effectively improving the overall convergence performance of the wireless FL system under the CAMU strategy.

\subsection{Complexity Analysis}

The computational complexity of the PPO algorithm in this chapter mainly stems from two aspects: the training updates of the Actor and Critic networks, and the computation of the convergence performance feedback from the environment given a resource configuration. Specifically, in each iteration, the agent first uses the Actor network to generate the action distribution of transmit power and local update frequency based on the current state $\boldsymbol{s}_n$, and samples a specific configuration; the forward propagation complexity is $\mathcal{O}(W)$, where $W$ is the total number of network parameters. Then, the environment computes the objective function value under the current action using the convergence upper bound formula from \textbf{Theorem 1}. As the environment is a static deterministic system, this process has a complexity of $\mathcal{O}(C)$, where $C$ is the number of clusters.

In the policy update phase, the algorithm samples $N$ trajectories and performs $\mathcal{K}$ inner-loop optimization steps. The forward and backward propagation complexity of each update round is $\mathcal{O}(\mathcal{K} \cdot N \cdot W)$. Therefore, the overall complexity per iteration is:
\begin{equation}
	\mathcal{O}(W + C + \mathcal{K} \cdot N \cdot W).
\end{equation}

If the algorithm runs for $E$ episodes, each containing $T_{{RL}}$ interaction steps, the total training complexity is:
\begin{equation}
	\mathcal{O}(E \cdot T_{{RL}} \cdot (W + C + \mathcal{K} \cdot N \cdot W)).
\end{equation}

In summary, since the environment is a static deterministic system, the cost of environment feedback is low. The overall complexity of the PPO algorithm is mainly influenced by the neural network training, offering good scalability and training efficiency. This makes it well-suited for the resource-coupled optimization problem proposed in this chapter.

\section{Simulation Results}

\subsection{Simulation Setup and Baselines}

\begin{figure}
	\centerline{\includegraphics[width=0.9\linewidth]{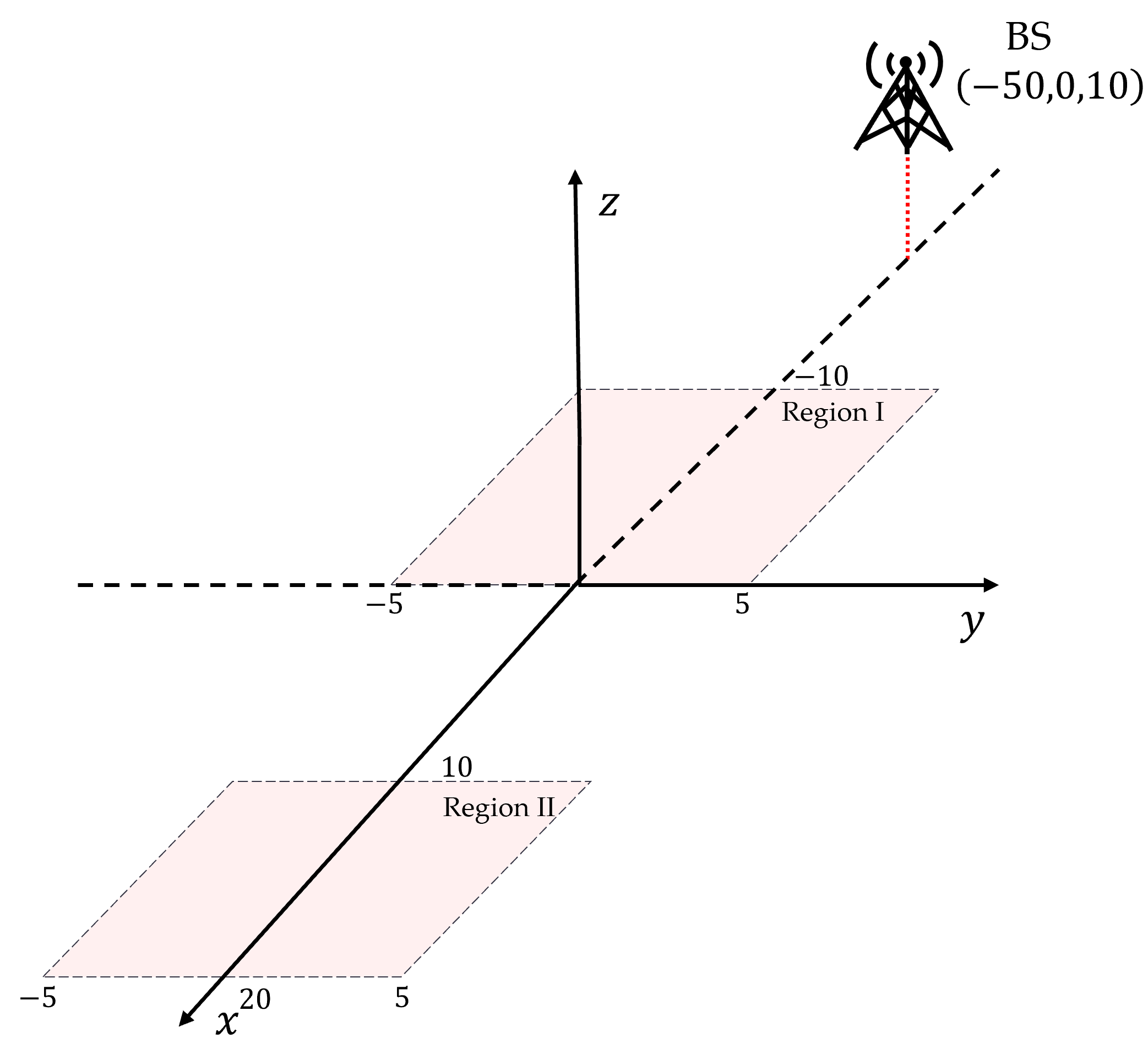}}
	\caption{Distribution of the participating devices.}
	\label{layout}
\end{figure}

The spatial relationships between all devices in the simulation are described using a Cartesian coordinate system, as shown in Fig. \ref{layout}. The BS, serving as the parameter server, is located at $(-50, 0, 10)\ \text{m}$. A total of $K=30$ devices are mainly distributed in two regions: 15 devices are randomly selected to be located in Region I, i.e., $\{(x, y, 0): -10 \leq x \leq 0,\ -5 \leq y \leq 5\}\ \text{m}$, and the other 15 devices are located in Region II, i.e., $\{(x, y, 0): 10 \leq x \leq 20,\ -5 \leq y \leq 5\}\ \text{m}$.

We use the mini-batch stochastic gradient descent algorithm for local model training, with a batch size set to $0.2$. The learning rate is set to $\lambda = 0.5 \times 10^{-3}$. The path loss in the wireless transmission system is given by:
\begin{equation}
	{PL}_{{DB}} = G_{{BS}} G_{{D}} \left( \frac{c}{4 \pi f_c d_{{DB}}} \right)^P,
\end{equation}
where the number of BS antennas is $N_a = 15$, the antenna gain is $G_{{BS}} = 5$ dBi, the device antenna gain is $G_{{D}} = 0$ dBi, the carrier frequency is $f_c = 915$ MHz, the path loss exponent is $P = 3.76$, $d_{{DB}}$ is the distance between the device and the BS, and $c$ is the speed of light. The noise power is set to $10^{-4}$ W.

The simulation uses a CNN network to train and test the MNIST and Fashion-MNIST datasets. The network consists of two $5 \times 5$ convolutional layers (each followed by a $2 \times 2$ max pooling layer), followed by a batch normalization layer, a fully connected layer with 50 units, a ReLU activation layer, and a softmax output layer. During training, the cross-entropy function is used as the loss function.

We first evaluates the improvement in aggregation efficiency provided by the proposed clustering mechanism under heterogeneous data and communication environments. It is then verified that the convergence advantages of the CAMU strategy in controlling local update frequencies and suppressing update bias. Finally, the ability of PPO-based joint resource optimization to enhance model performance under energy-limited conditions is analyzed.

\subsection{Effectiveness of the Aggregation Strategy}

This section verifies the effectiveness of the proposed device clustering strategy. The simulation setup includes the following configurations:

\textbf{Configuration 1}: Aggregation using the clustering strategy proposed in this paper.

\textbf{Configuration 2}: Heterogeneous devices are not clustered, but aggregation weights are calculated using Wasserstein distance (i.e., formula \eqref{Gc}).

Additionally, the simulation compares with the widely used FedAvg algorithm (denoted as \textbf{Benchmark 1}), where devices directly upload their local model parameters without clustering, and aggregation weights are determined by the data size.

\begin{figure*}[htbp]
    \centering
    \begin{minipage}{0.18\linewidth}
        \centering
        \quad \small{\small{24 IID+\textcolor{red}{6 Non-IID}}}
    \end{minipage}
    \begin{minipage}{0.18\linewidth}
        \centering
        \quad \small{\small{18 IID+\textcolor{red}{12 Non-IID}}}
    \end{minipage}
    \begin{minipage}{0.18\linewidth}
        \centering
        \quad \small{\small{12 IID+\textcolor{red}{18 Non-IID}}}
    \end{minipage}
    \begin{minipage}{0.18\linewidth}
        \centering
        \quad \small{\small{6 IID+\textcolor{red}{24 Non-IID}}}
    \end{minipage}
    \begin{minipage}{0.18\linewidth}
        \centering
        \quad \small{\small{0 IID+\textcolor{red}{30 Non-IID}}}
    \end{minipage}\\[3pt]
    
    \begin{minipage}{0.18\linewidth}
        \centering
        \includegraphics[width=\linewidth]{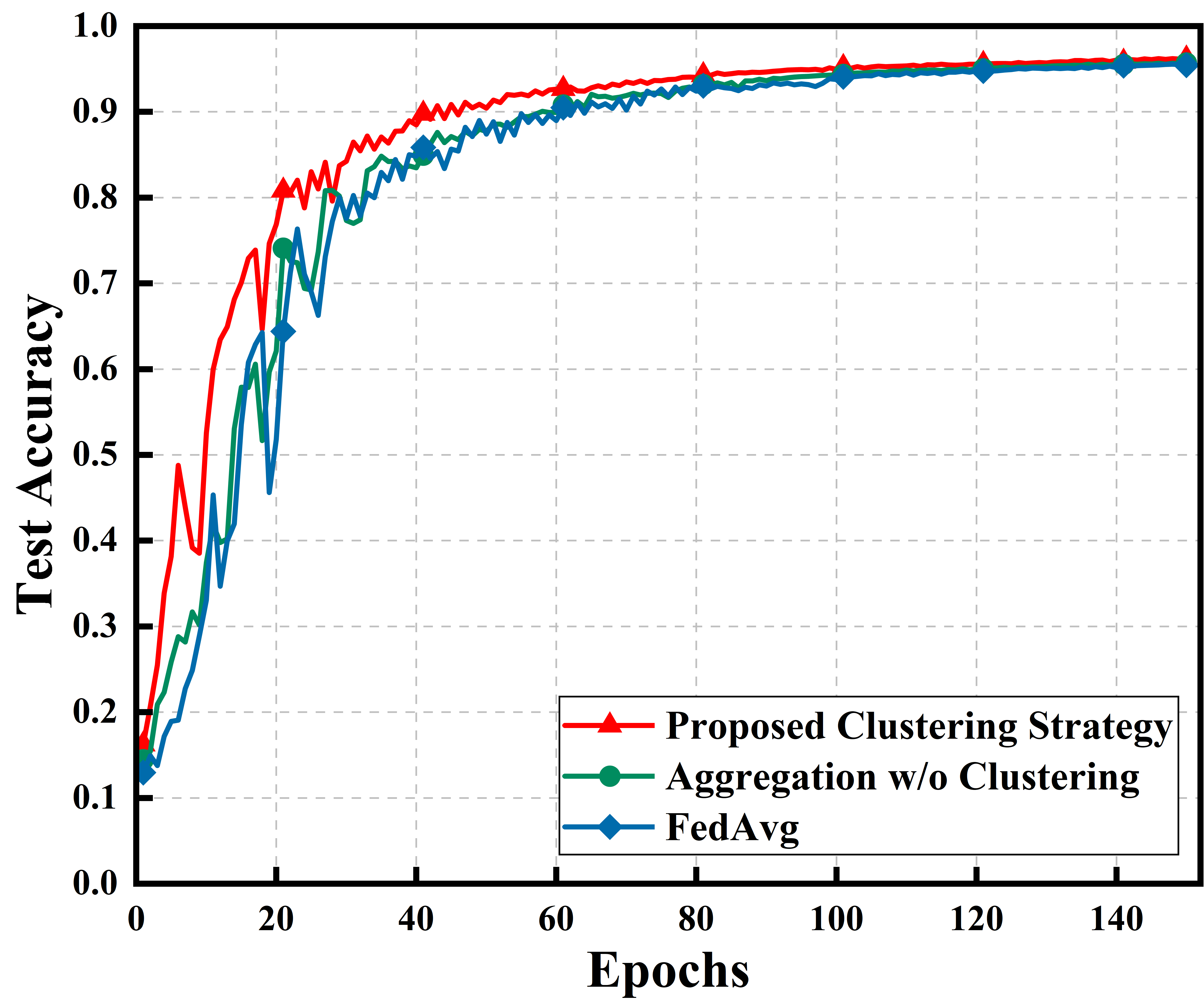}
    \end{minipage}
    \begin{minipage}{0.18\linewidth}
        \centering
        \includegraphics[width=\linewidth]{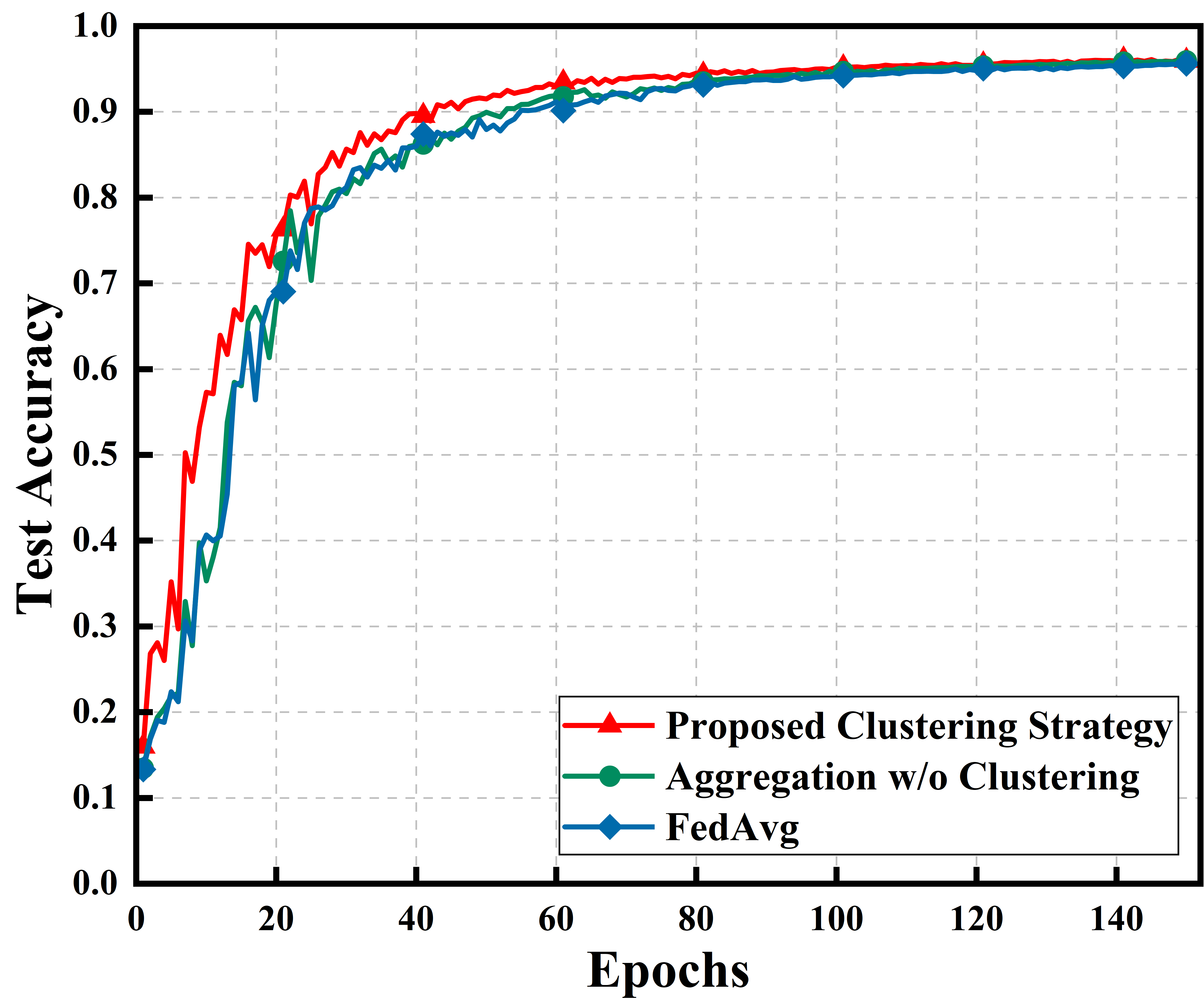}
    \end{minipage}
    \begin{minipage}{0.18\linewidth}
        \centering
        \includegraphics[width=\linewidth]{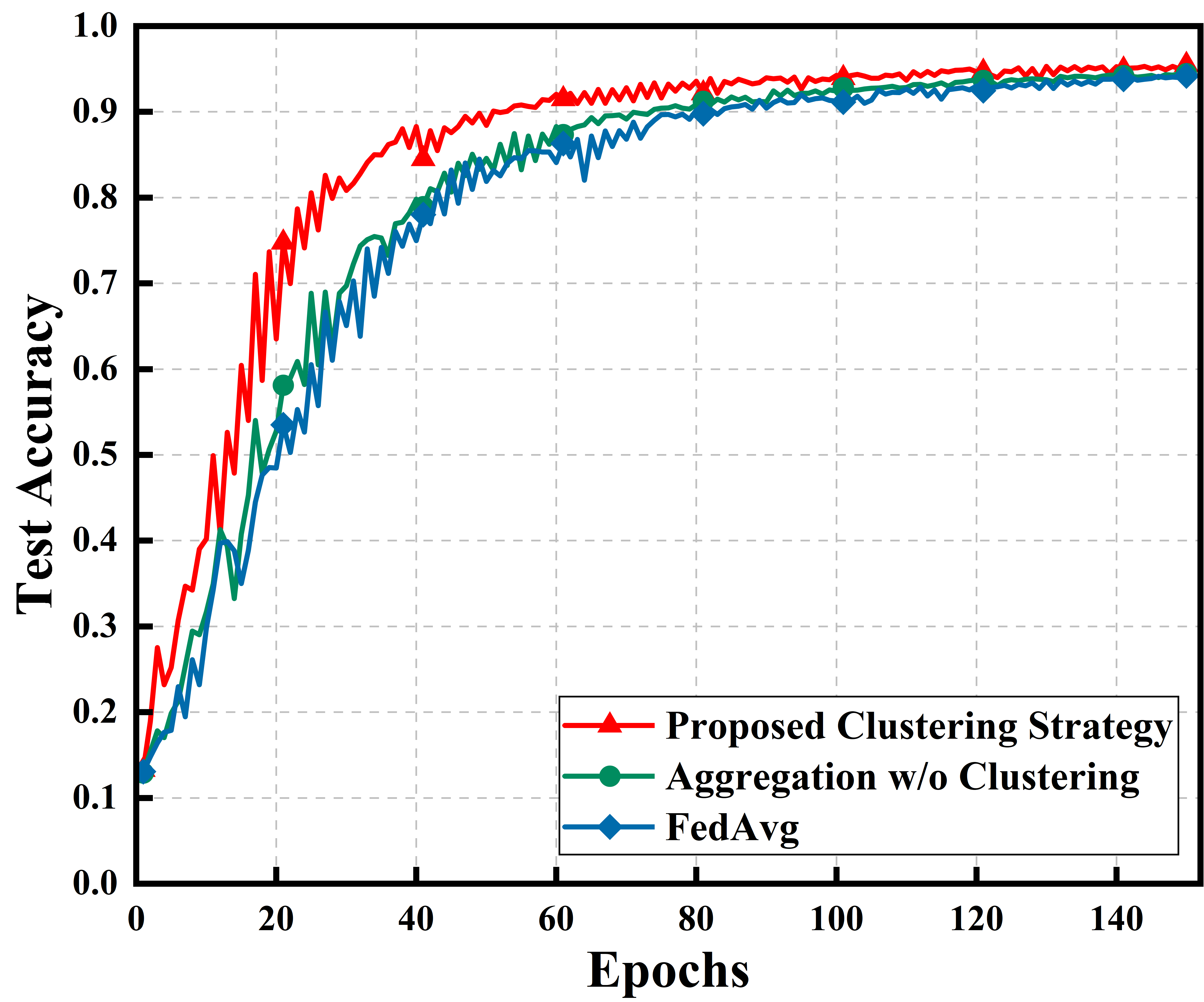}
    \end{minipage}
    \begin{minipage}{0.18\linewidth}
        \centering
        \includegraphics[width=\linewidth]{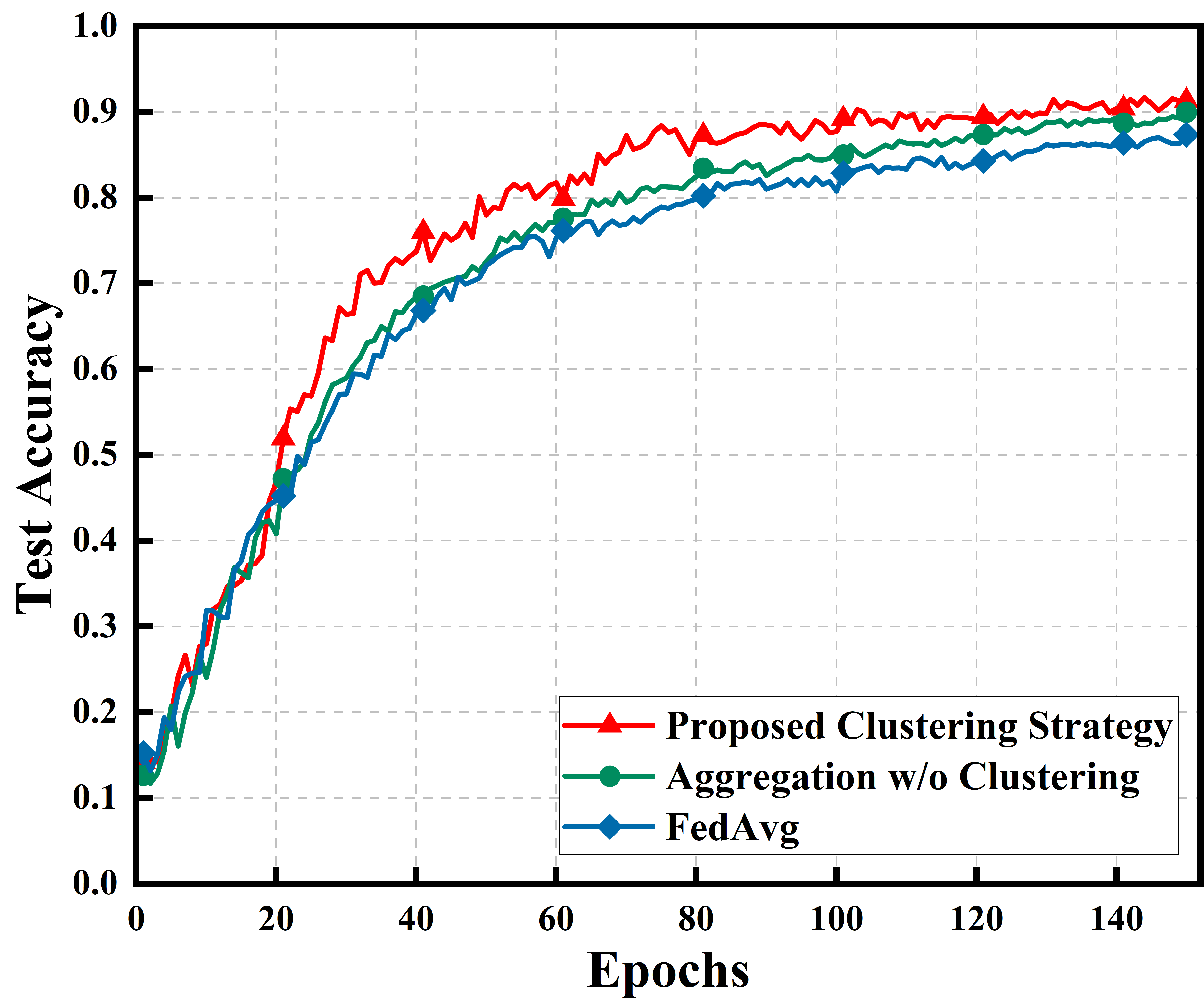}
    \end{minipage}
    \begin{minipage}{0.18\linewidth}
        \centering
        \includegraphics[width=\linewidth]{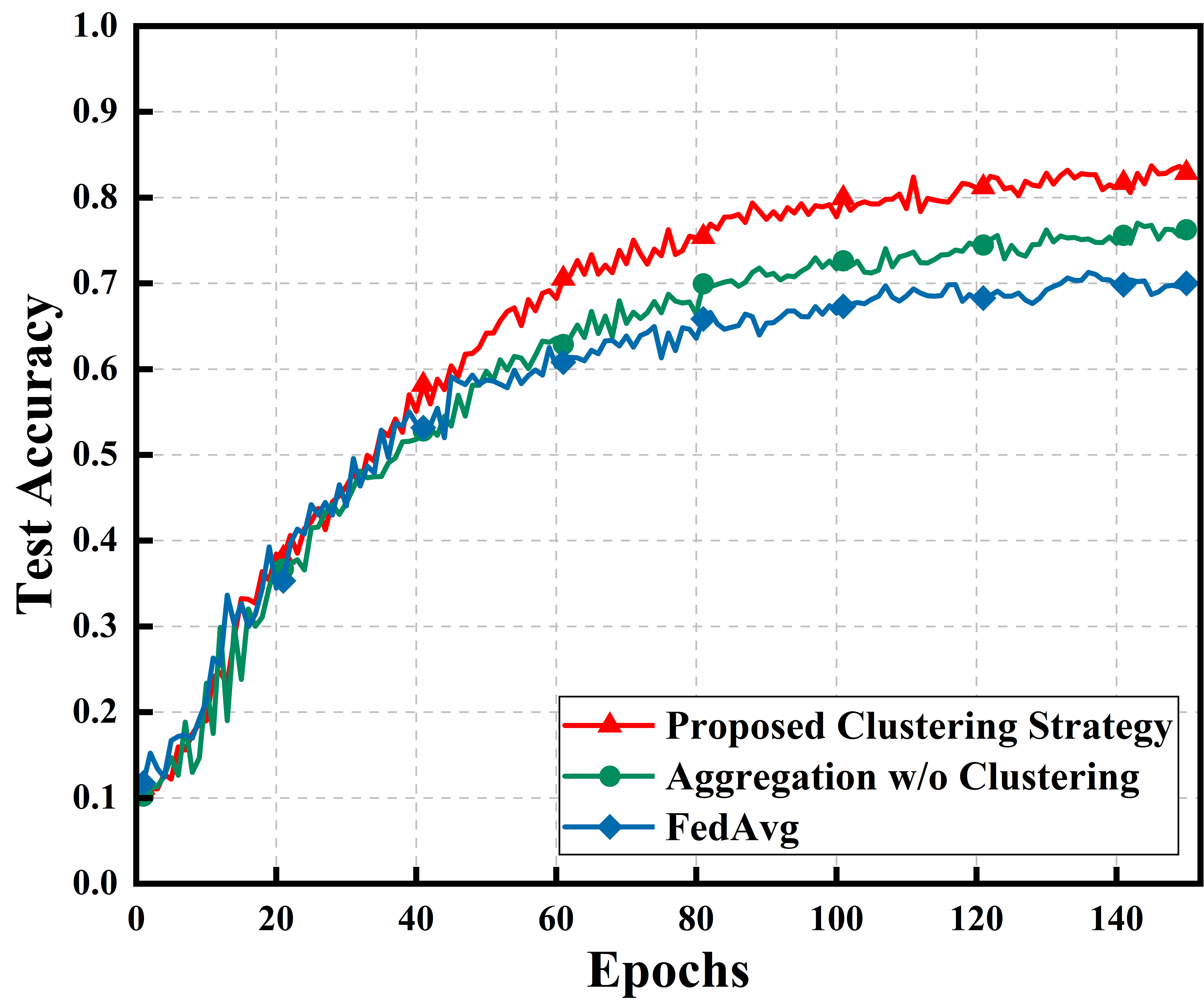}
    \end{minipage}
    \\[5pt] \parbox{0.5\linewidth}{\centering (a) MNIST}\\[5pt]

    \begin{minipage}{0.18\linewidth}
        \centering
        \includegraphics[width=\linewidth]{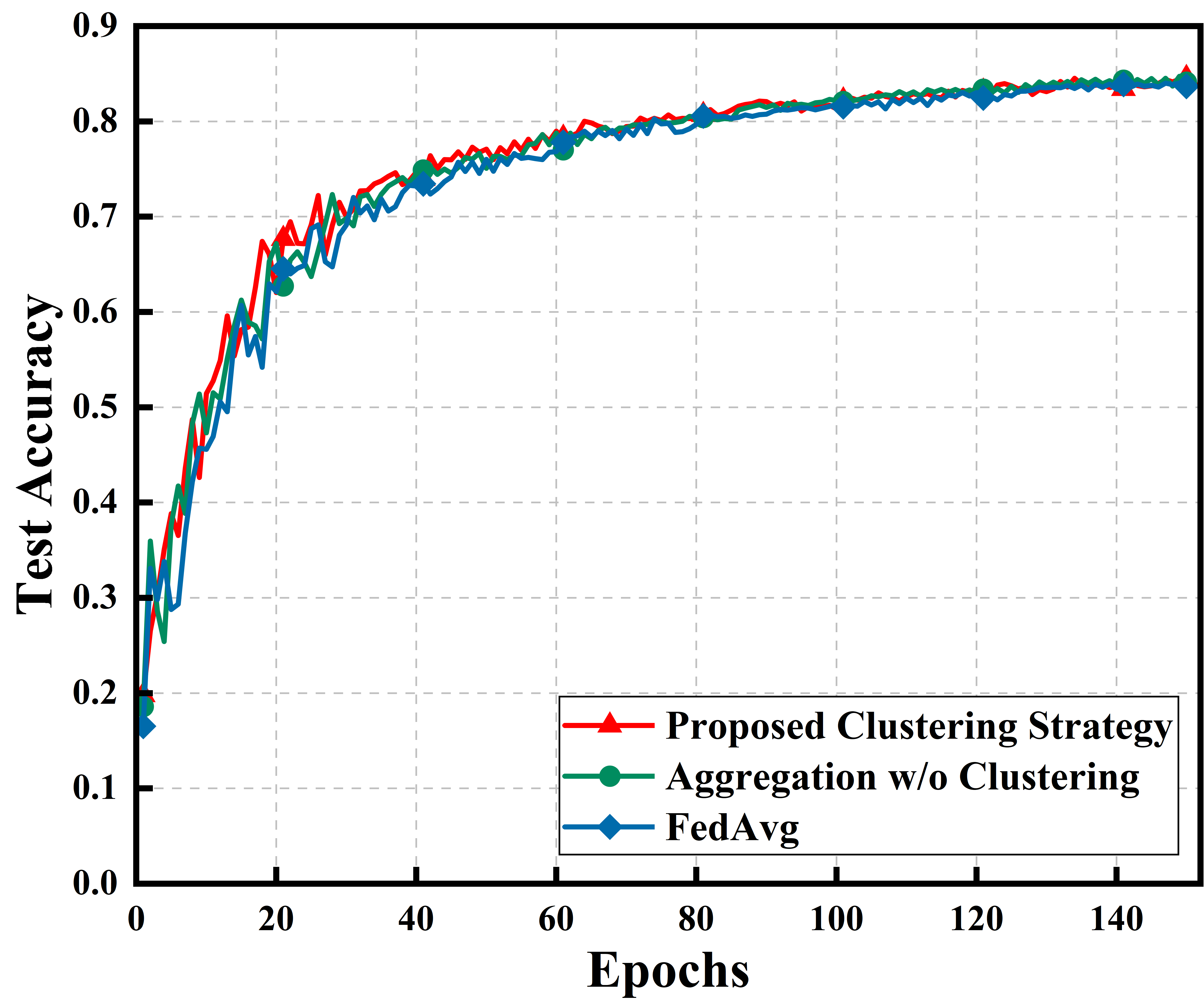}
    \end{minipage}
    \begin{minipage}{0.18\linewidth}
        \centering
        \includegraphics[width=\linewidth]{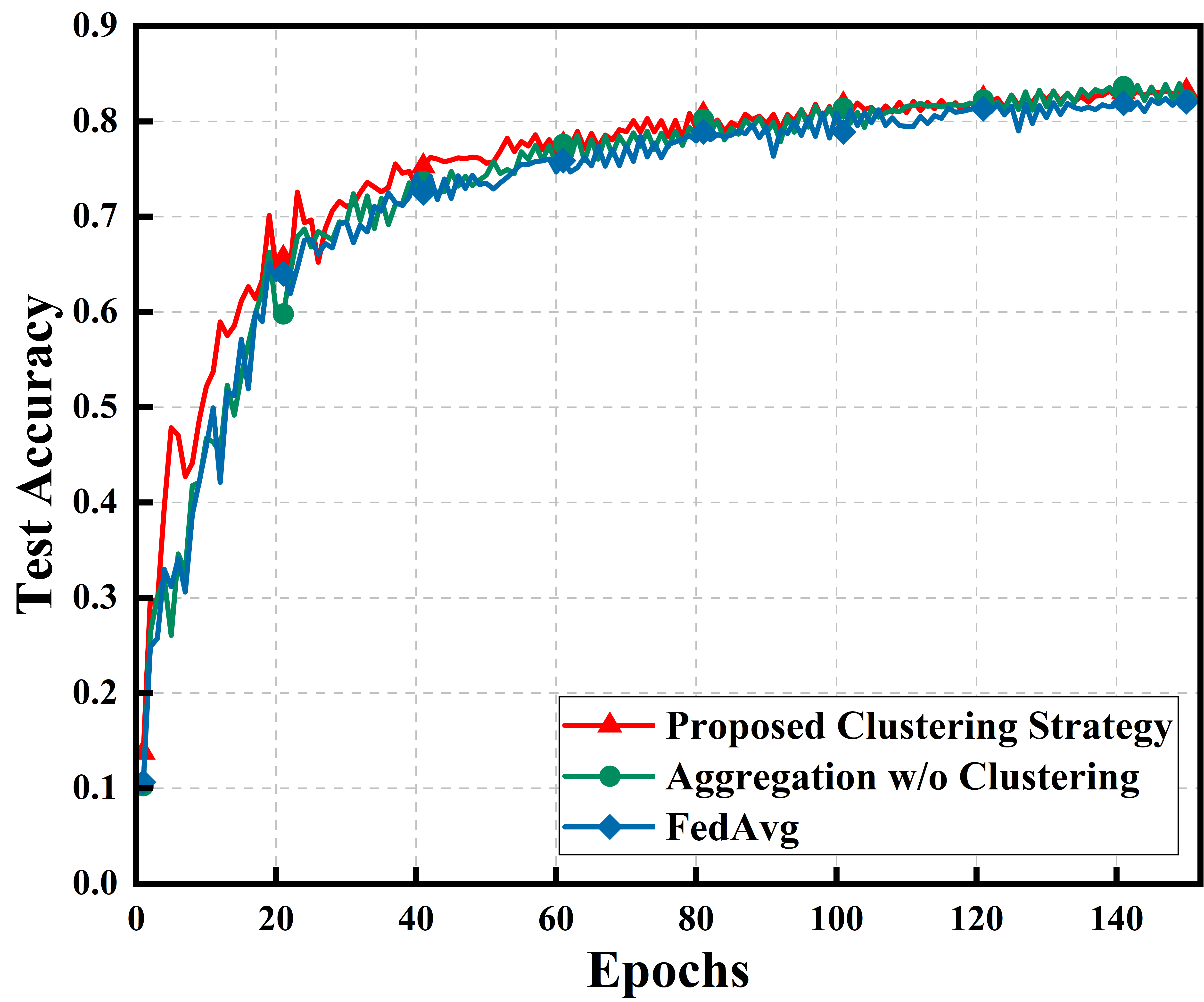}
    \end{minipage}
    \begin{minipage}{0.18\linewidth}
        \centering
        \includegraphics[width=\linewidth]{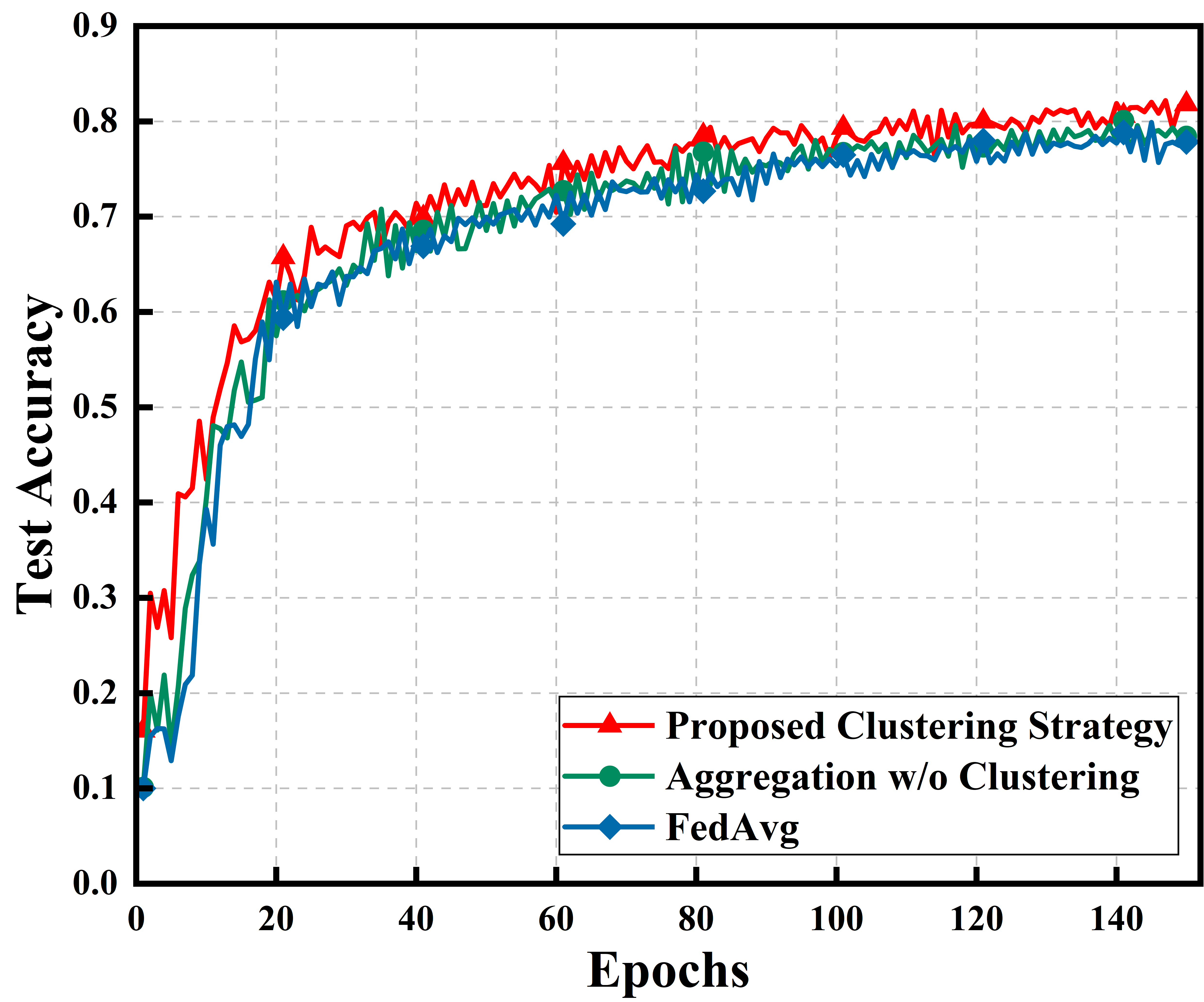}
    \end{minipage}
    \begin{minipage}{0.18\linewidth}
        \centering
        \includegraphics[width=\linewidth]{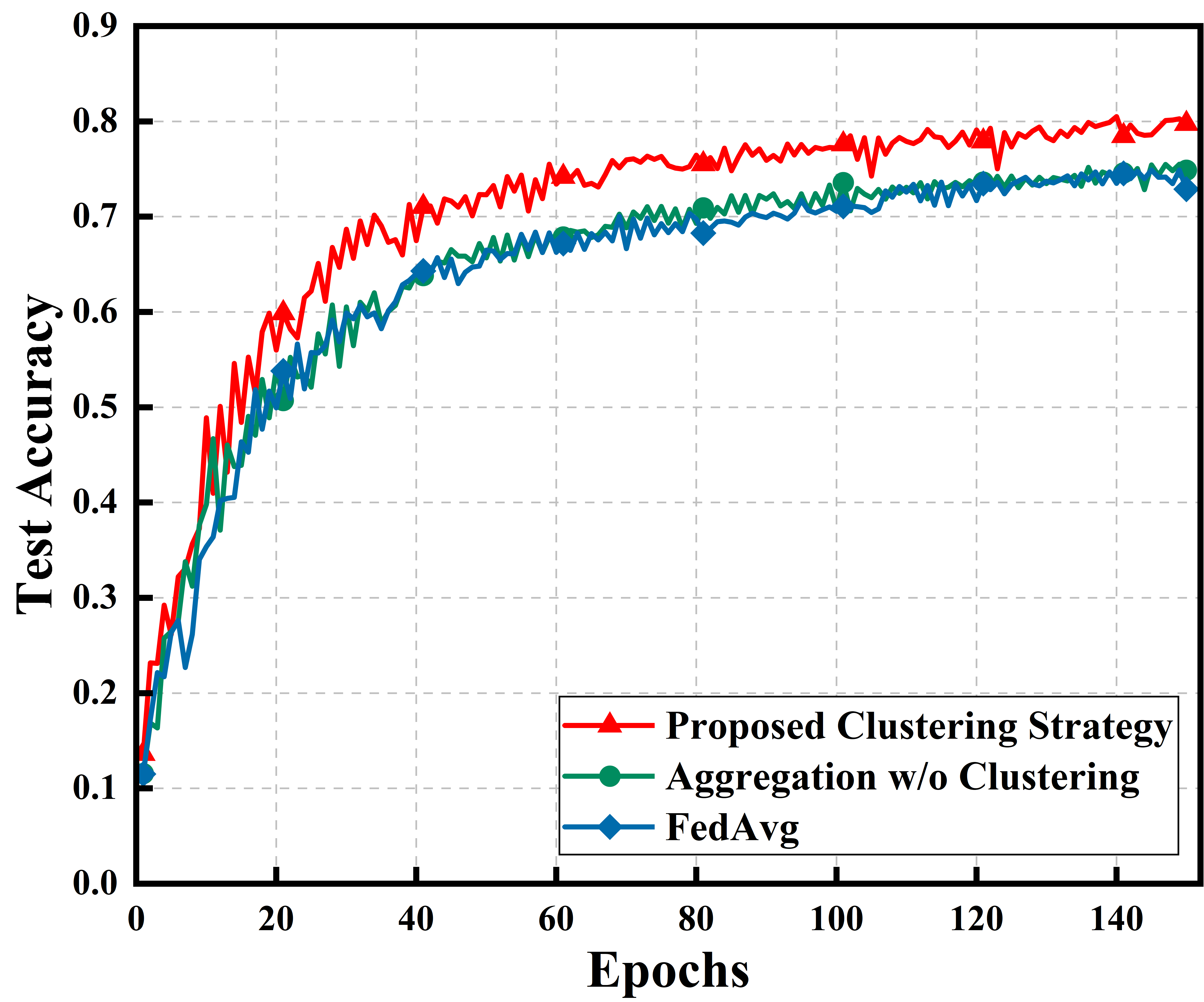}
    \end{minipage}
    \begin{minipage}{0.18\linewidth}
        \centering
        \includegraphics[width=\linewidth]{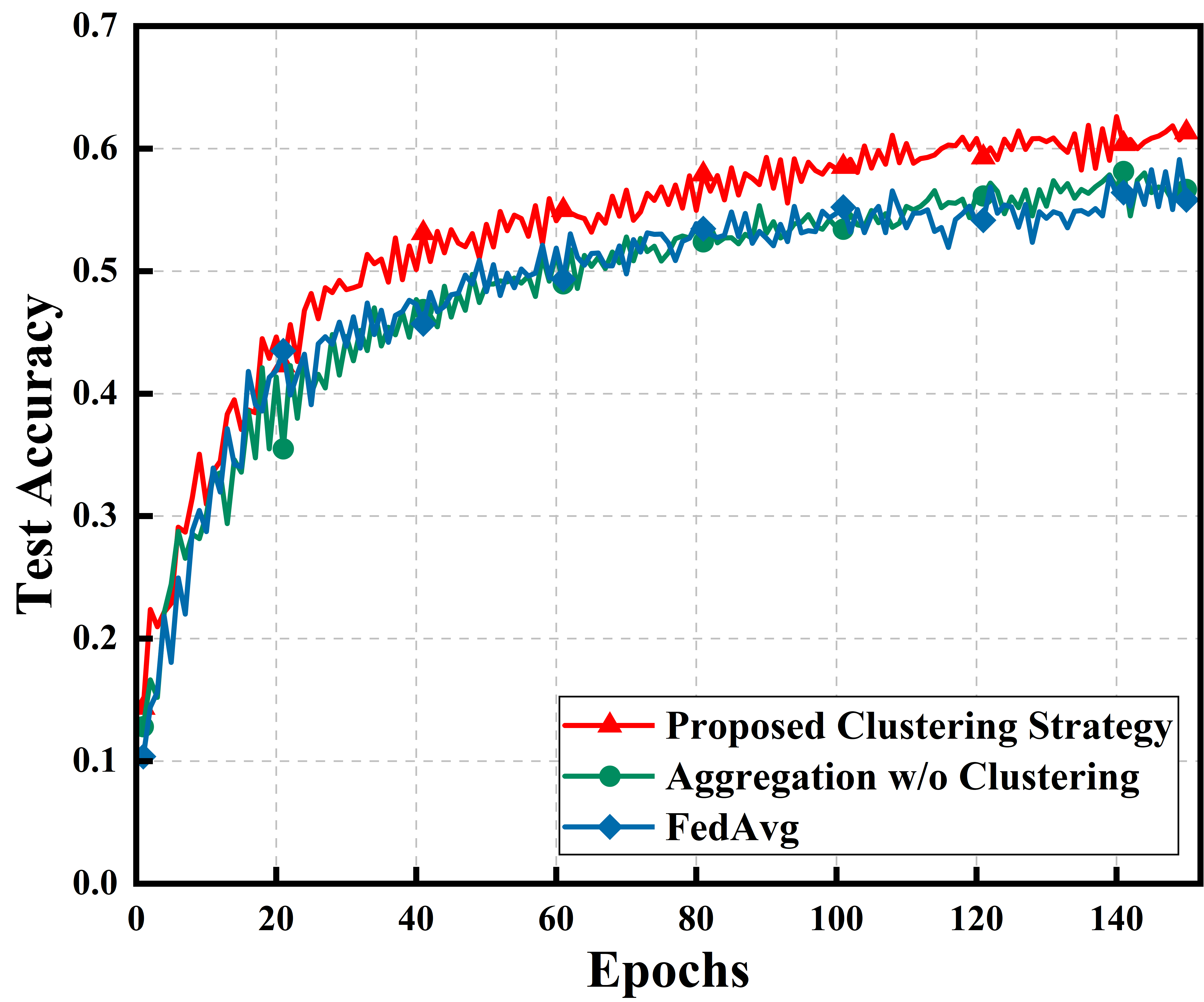}
    \end{minipage}
     \\[5pt] \parbox{0.5\linewidth}{\centering (b) Fashion-MNIST}

    \caption{FL convergence performance based on each aggregation scheme under different degree of data heterogeneity.}
    \label{exp1}
\end{figure*}

The simulations are conducted on both the MNIST and Fashion-MNIST datasets. For each dataset, different proportions of devices with Non-IID data are set to observe the performance of the clustering strategy under various heterogeneity conditions. The results are shown in Fig. \ref{exp1}.

As seen in the figure, from left to right, the proportion of devices with Non-IID data gradually increases, indicating an increase in data heterogeneity. The comparison results show that, for both datasets, the convergence performance of {Configuration 2} is superior to {Benchmark 1}. Specifically, on the MNIST dataset, the performance improvement of {Configuration 2} compared to {Benchmark 1} increases from 0.255\% to 6.262\% as the heterogeneity level rises; on the Fashion-MNIST dataset, the performance improvement ranges from 0.431\% to 0.858\%. This performance improvement is due to the personalized aggregation weights computed based on Wasserstein distance, which effectively mitigates the degradation in convergence performance caused by data heterogeneity.

Furthermore, it reveals that the convergence performance of {Configuration 1} is consistently better than that of {Configuration 2} across all settings. Specifically, the aggregation performance improvement on the MNIST dataset increases from 0.507\% to 6.615\%, while the performance improvement on the Fashion-MNIST dataset ranges from 0.427\% to 4.645\%. This result validates the effectiveness of the clustering strategy proposed in this paper under heterogeneous conditions. It is noteworthy that as data heterogeneity increases, the advantages of the proposed clustering strategy become more pronounced, indicating that reducing the number of Non-IID individuals through clustering, under the premise of treating clusters as new aggregation entities, helps enhance the overall performance of FL. Although the improvement is not as significant as that of the clustering method proposed in the previous work \cite{sun2025dual}, which mitigates heterogeneity through communication consistency, it provides a structural foundation for the consistency of cluster-level actions in the subsequent CAMU mechanism.

\subsection{Superiority of the CAMU Strategy}

The simulation in this section aims to verify the superiority of the proposed CAMU strategy. As previously noted, configuring the local update frequency solely based on cluster computational capacity may cause frequent updates of low-contribution clusters under data heterogeneity, degrading aggregation performance.
Therefore, clusters with low contribution should be restricted to a single local update per global round. To evaluate the superiority of this strategy, the following configurations are considered in the simulation:

\textbf{Configuration 3}: The CAMU strategy based on different cluster contribution thresholds, with thresholds set at $0.5 \times 10^4$, $1.0 \times 10^4$, and $1.5 \times 10^4$.

\textbf{Configuration 4}: Each cluster performs only one local update in each global aggregation round.

\begin{figure}
    \begin{subfigure}{\columnwidth} 
        \centering
        \includegraphics[width=0.98\columnwidth]{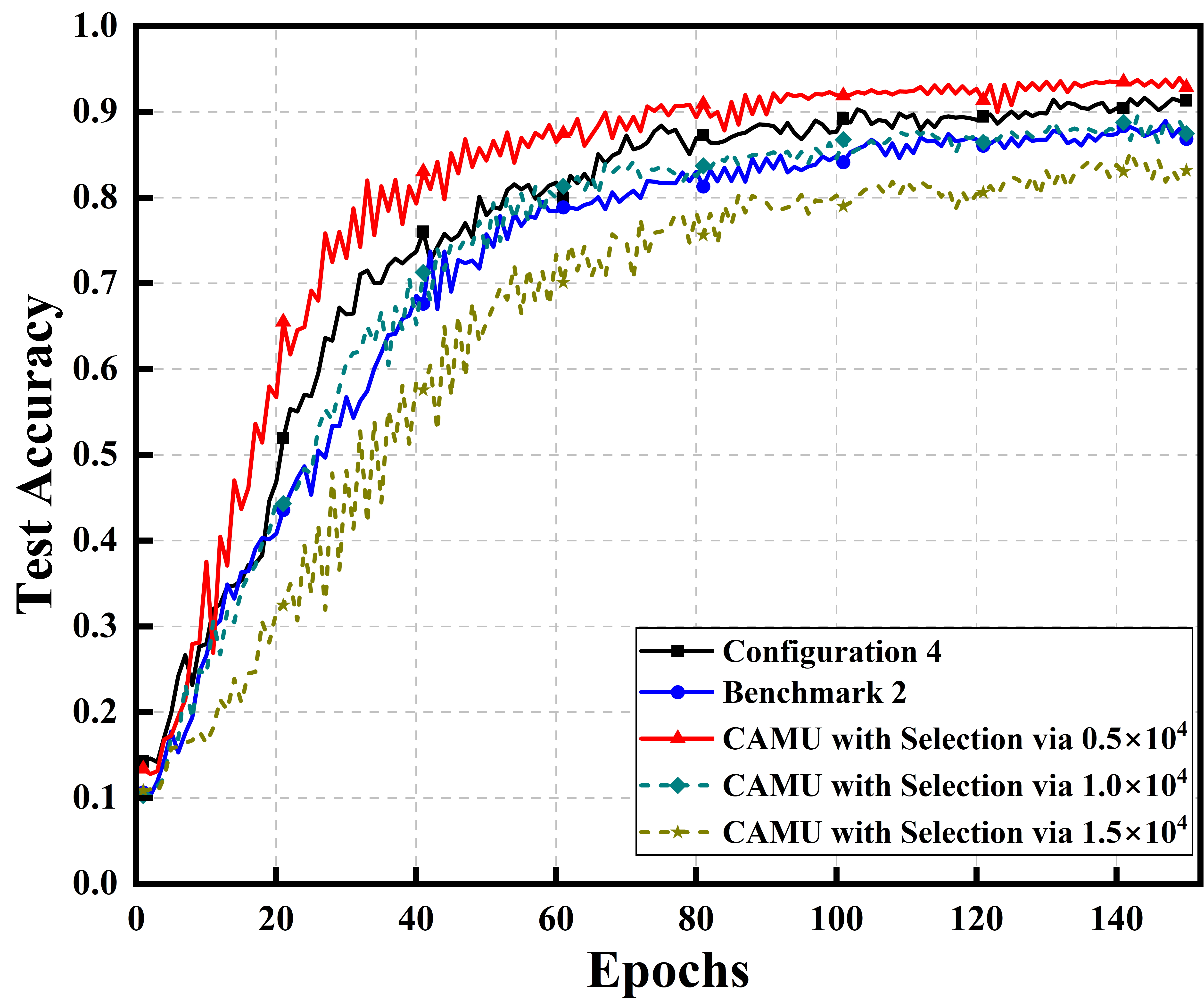} 
        \caption{MNIST}
        \includegraphics[width=0.98\columnwidth]{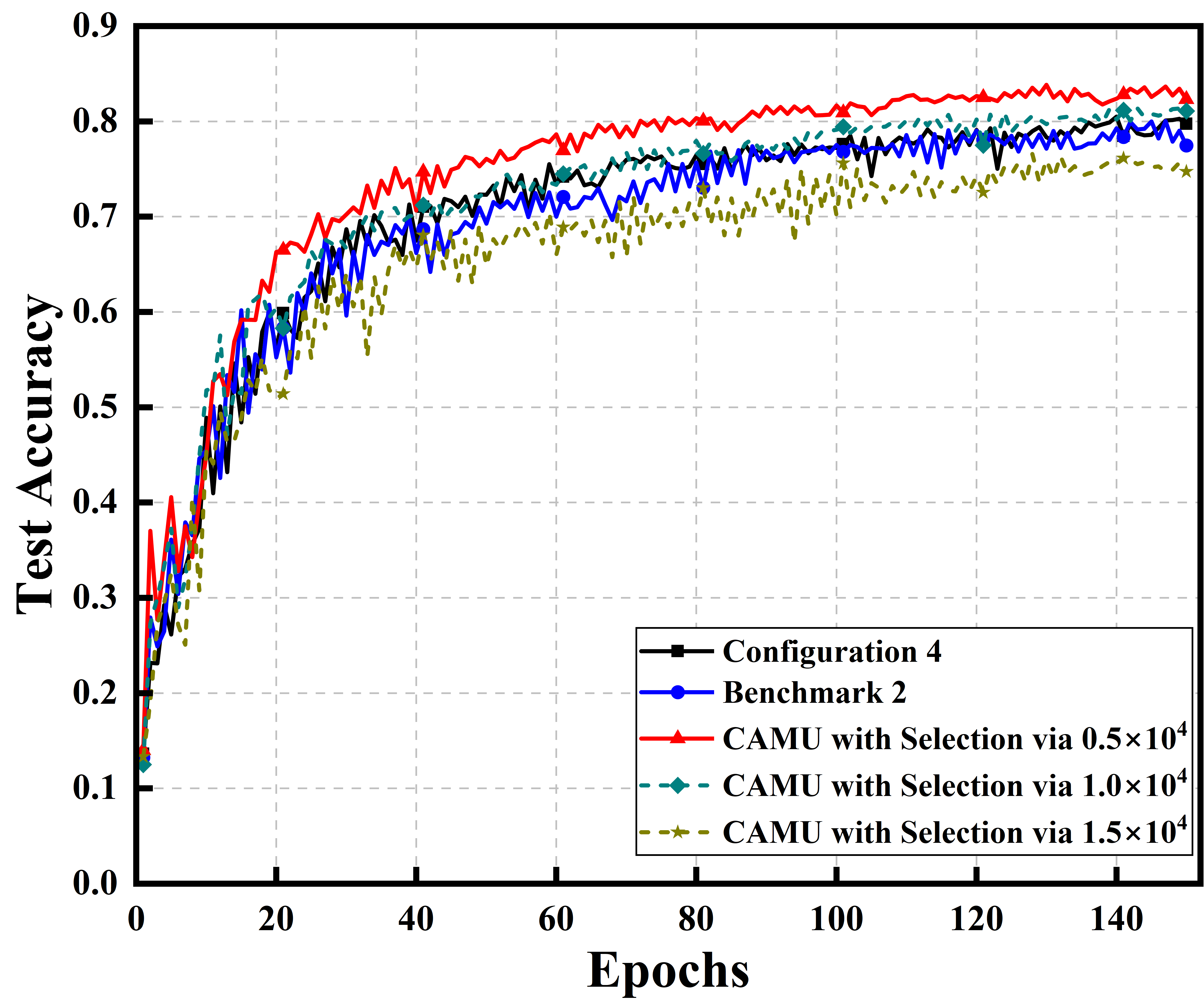} 
        \caption{Fashion-MNIST}
    \end{subfigure}
    \caption{Comparison of aggregation performance between local multi-round updates with clusters as basic update units and single-round updates.}
\label{exp2}
\end{figure}

Additionally, based on the work of \cite{zhou2020petrel}, this simulation implements a local multi-round update scheme with clusters as the update unit, where the local update frequency is determined by cluster computational capacity to minimize system waiting time, without considering cluster contribution (denoted as \textbf{Benchmark 2}). The simulation results are shown in Fig. \ref{exp2}.


From the comparison results of the two datasets, it can be seen that the local multi-round update strategy generally helps improve the convergence performance of FL. However, further comparison between Benchmark 2 and Configuration 4 shows that, when all devices have data with heterogeneous distributions, i.e., when the system heterogeneity is high, the aggregation performance of Benchmark 2 is 4.468\% lower than Configuration 4 on the MNIST dataset and 2.252\% lower on the Fashion-MNIST dataset. This result indicates that, in highly heterogeneous scenarios, if the update units are not selected based on data contribution, frequent participation of low-contribution clusters in local multi-round updates may lead to significant bias accumulation, negatively affecting the global update, as discussed in Section III-B.

Furthermore, a comparison of Configuration 3 with Benchmark 2 and Configuration 4 reveals that under heterogeneous conditions, the CAMU strategy significantly improves convergence performance by filtering low-contribution clusters. Compared to Configuration 4, the CAMU strategy improves performance by up to 1.527\% and 2.573\% on the MNIST and Fashion-MNIST datasets, respectively. Compared to Benchmark 2, the maximum improvements are 5.995\% and 4.825\%, respectively. Additionally, as the cluster contribution threshold increases, the convergence performance gradually decreases. This is because higher thresholds allow more low-contribution clusters to participate in local multi-round updates, thereby increasing the risk of bias accumulation.

The results clearly demonstrate that, in heterogeneous environments, filtering clusters based on data contribution during local multi-round updates is essential, validating the effectiveness and superiority of the CAMU strategy in enhancing FL performance.

\subsection{CAMU-Based Resource Optimization}

Based on the proposed CAMU strategy, this section jointly optimizes the clustered local update frequency and transmit power under resource-constrained conditions using the PPO algorithm (denoted as \textbf{Configuration 5}). The simulation first assigns the same transmit power (0.5W) to each cluster in Benchmark 2 in the actual communication environment. To ensure fairness, the total energy consumed in the Benchmark 2 scheme is used as the energy constraint for PPO optimization, where the energy includes both the computational energy consumed by local updates and the communication energy required for transmitting model parameters from the cluster leader to the BS. This setup ensures a fair comparison between the two schemes under the same resource conditions.

\begin{figure}
    \begin{subfigure}{\columnwidth} 
        \centering
        \includegraphics[width=0.98\columnwidth]{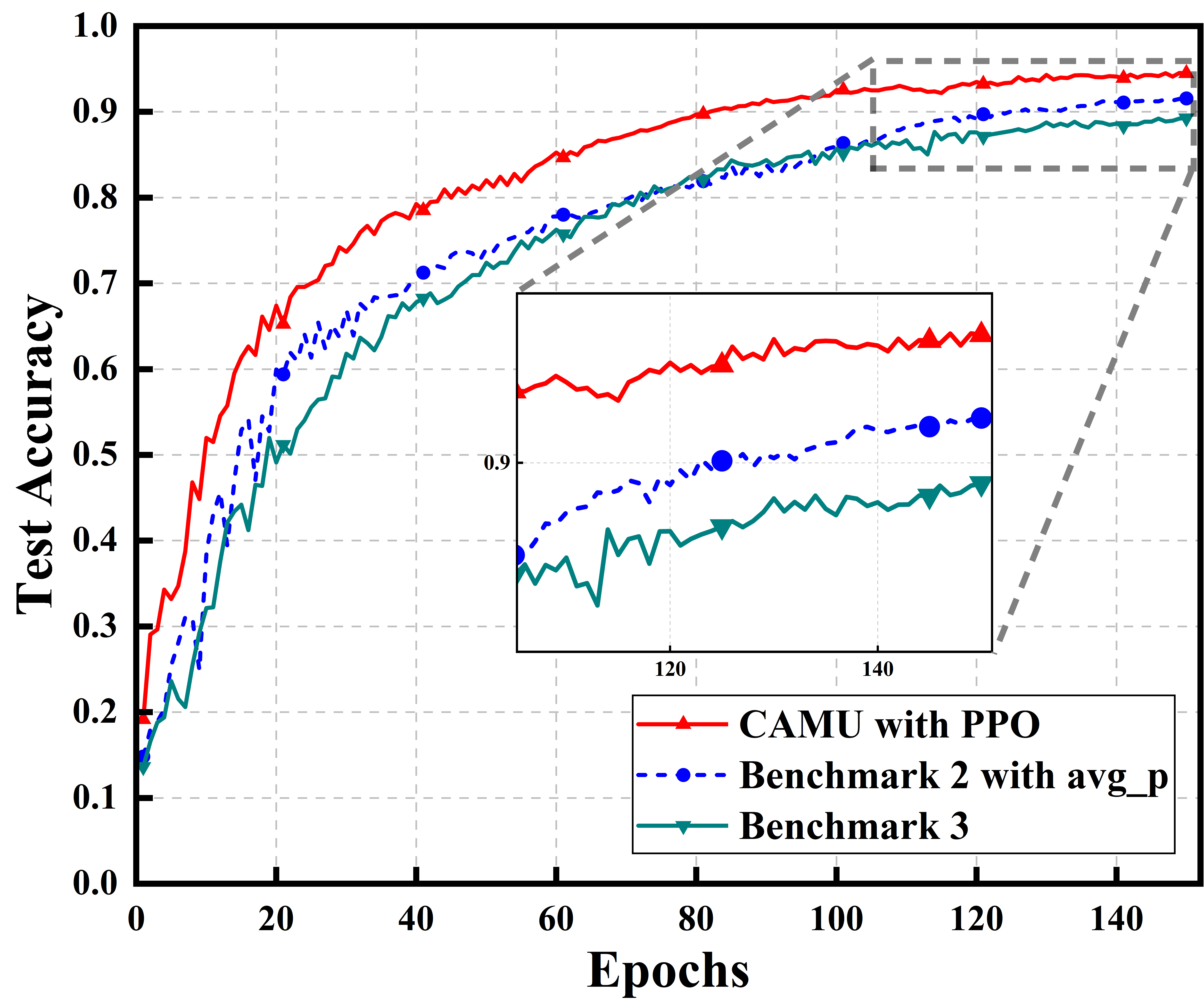} 
        \caption{MNIST}
        \includegraphics[width=0.98\columnwidth]{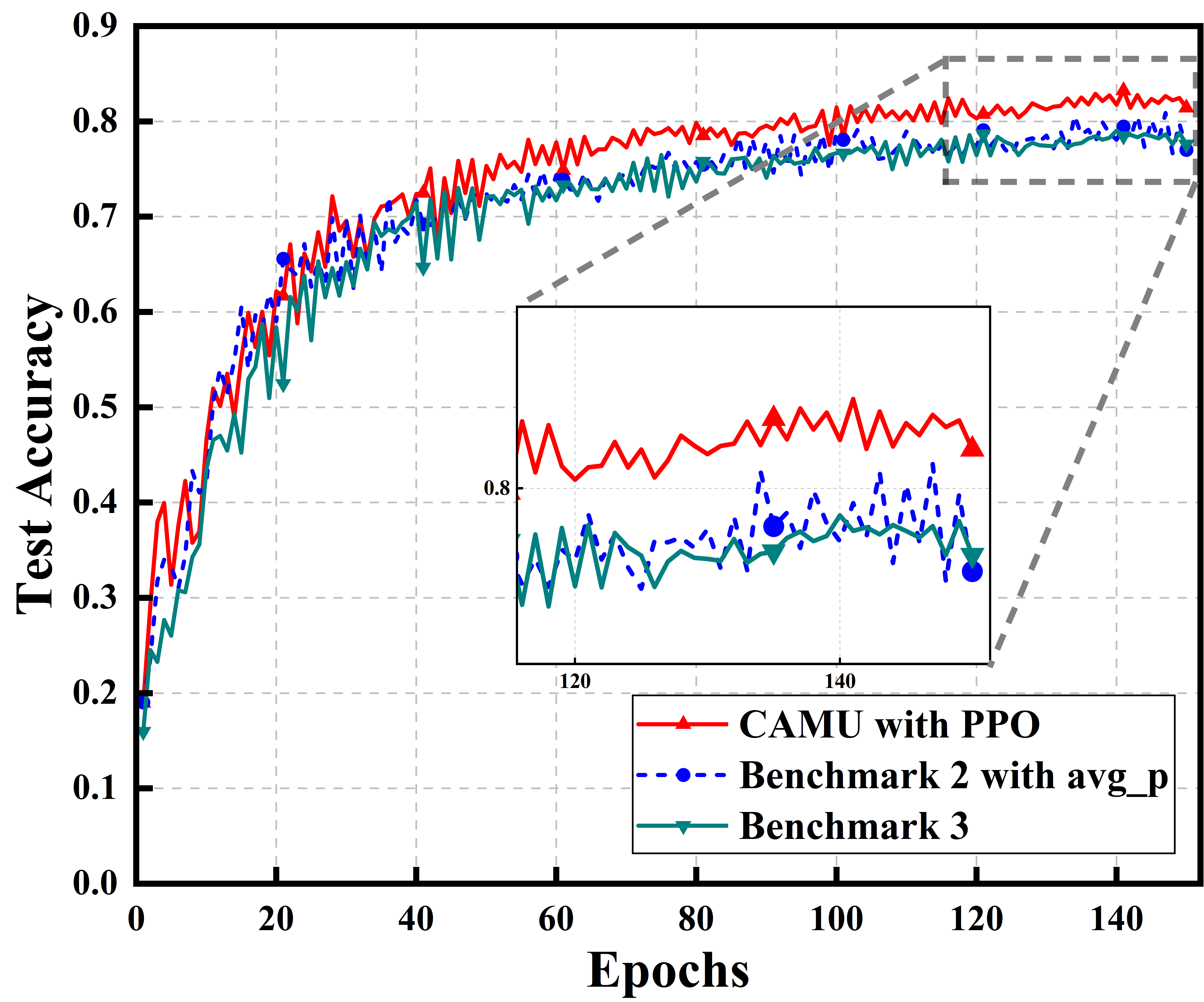} 
        \caption{Fashion-MNIST}
    \end{subfigure}
    \caption{Performance comparison of PPO-based joint optimization schemes based on CAMU strategy under resource-constrained conditions.}
\label{exp3}
\end{figure}

The simulation compares the scheme in the existing advanced work \cite{zeng2022heterogeneous}, which only optimizes the number of local iterations (denoted as \textbf{Benchmark 3}). To ensure fairness, the update units in Benchmark 3 are set as clusters. The experimental results are shown in Fig. \ref{exp3}.

As shown in the figure, under actual communication conditions with limited system resources, the CAMU-based PPO joint optimization strategy outperforms Benchmark 2, improving convergence accuracy by 2.910\% on the MNIST dataset and 4.407\% on the Fashion-MNIST dataset. This highlights the critical importance of balanced allocation of computational and communication resources under energy constraints. Compared to the fixed transmit power and computation-driven local iteration strategy of Benchmark 2, the CAMU strategy achieves superior FL performance within the same resource budget through joint optimization of local computation and communication. These results confirm the effectiveness of the proposed optimization scheme in resource-constrained scenarios and underscore the value of integrating intelligent optimization strategies at the system design stage to enhance overall efficiency.

In addition, compared with Benchmark 3, the proposed joint optimization algorithm improves convergence accuracy by 5.198\% on the MNIST dataset and 3.820\% on the Fashion-MNIST dataset. This further demonstrates the importance of jointly considering computational and communication capabilities in practical heterogeneous environments. While computational heterogeneity significantly impacts convergence efficiency, in real-world communication-heterogeneous scenarios, proper allocation of communication resources is equally critical. Overemphasizing computational capacity may lead to insufficient communication power, resulting in transmission error and degraded global convergence. Therefore, the proposed joint optimization of local update frequency and communication resources offers clear performance advantages in heterogeneous, resource-constrained environments.

In summary, the CAMU-based PPO joint optimization strategy proposed in this paper achieves superior convergence performance under resource constraints, demonstrating strong practical potential for heterogeneous wireless FL systems.

\section{Conclusion}

This paper addresses the challenges of data heterogeneity, communication heterogeneity, and resource constraints in wireless FL by proposing a CAMU strategy based on device clustering. Additionally, the PPO algorithm is employed to optimize the local update frequency and transmit power configuration within clusters under energy constraints. A cluster contribution filtering mechanism is introduced to effectively suppress the accumulation of local update bias.
Theoretical analysis and simulation results demonstrate that the proposed method significantly improves model convergence accuracy while ensuring system stability, with particularly strong performance in scenarios characterized by data heterogeneity and resource constraints.

\section{Appendix I}

For hierarchical FL, under practical communication environments, it is assumed that intra-cluster transmission introduces negligible communication error due to similar communication capacities and short distances between members and the group leader. Therefore, communication error is introduced only during the transmission from the group leader to the BS. Thus, we have:
\begin{equation}
	\begin{split}
		&\boldsymbol{w}^{[t+1]} \\
        &= \sum_{c=1}^{C} G_c \left( \boldsymbol{w}_c^{[t,N_c]} + \frac{\sigma_n}{p_c \|\boldsymbol{h}_c\|} \right) \\
		&= \boldsymbol{w}^{[t]} - \lambda \sum_{c=1}^{C} G_c \sum_{n=0}^{N_c - 1} \sum_{k=1}^{K_c} G_{k,c} \nabla F_{k,c}\left(\boldsymbol{w}_c^{[t,n]}\right) \\
        & \quad + \sum_{c=1}^{C} G_c \frac{\sigma_n}{p_c \|\boldsymbol{h}_c\|},
	\end{split}
\end{equation}
i.e.,
\begin{equation}
	\begin{split}
		&\boldsymbol{w}^{[t+1]} - \boldsymbol{w}^{[t]} \\
        &= - \lambda \sum_{c=1}^{C} G_c \sum_{n=0}^{N_c - 1} \sum_{k=1}^{K_c} G_{k,c} \nabla F_{k,c}\left(\boldsymbol{w}_c^{[t,n]}\right) \\
		&\quad + \sum_{c=1}^{C} G_c \frac{\sigma_n}{p_c \|\boldsymbol{h}_c\|}.
	\end{split}
\end{equation}

By performing a Taylor expansion of $F(\boldsymbol{w}^{[t+1]})$ and taking the expectation, we obtain:
\begin{equation}
	\begin{split}
		&\mathbb{E}[F(\boldsymbol{w}^{[t+1]})]\\
		& \leq \mathbb{E}[F(\boldsymbol{w}^{[t]})] + \mathbb{E}[(\boldsymbol{w}^{[t+1]} - \boldsymbol{w}^{[t]})^{\top} \nabla F(\boldsymbol{w}^{[t]})] \\
		& \quad + \frac{L}{2} \mathbb{E}\left[\left\| \boldsymbol{w}^{[t+1]} - \boldsymbol{w}^{[t]} \right\|^2\right].
	\end{split}
\end{equation}
where
\begin{equation}
	\begin{split}
		&\mathbb{E}\left[\left\| \boldsymbol{w}^{[t+1]} - \boldsymbol{w}^{[t]} \right\|^2\right]\\
		&= \lambda^2 \mathbb{E}\left[\left\| \sum_{c=1}^{C} G_c \sum_{n=0}^{N_c - 1} \sum_{k=1}^{K_c} G_{k,c} \nabla F_{k,c}(\boldsymbol{w}_c^{[t,n]}) \right\|^2 \right] \\
		& \quad + \mathbb{E}\left[\left\| \sum_{c=1}^{C} G_c \frac{\sigma_n}{p_c \|\boldsymbol{h}_c\|} \right\|^2 \right].
	\end{split}
\end{equation}

According to the Cauchy-Schwarz inequality, we have:
\begin{equation}
	\begin{split}
		&\mathbb{E}\left[\left\| \boldsymbol{w}^{[t+1]} - \boldsymbol{w}^{[t]} \right\|^2\right]\\
		& \leq \lambda^2 \sum_{c=1}^{C} G_c^2 \sum_{n=0}^{N_c - 1} \sum_{k=1}^{K_c} G_{k,c}^2 \mathbb{E}\left[\left\| \nabla F_{k,c}(\boldsymbol{w}_c^{[t,n]}) \right\|^2 \right] \\
		& \quad + \sum_{c=1}^{C} G_c^2 \frac{\sigma_n^2}{p_c^2 \|\boldsymbol{h}_c\|^2} \\
		& = \lambda^2 \sum_{c=1}^{C} \left( \sum_{k=1}^{K_c} G_{k,c}^2 \right) \left(G_c^2 \delta_c^2\right) \left( \sum_{n=0}^{N_c - 1} \left\| \nabla F_c(\boldsymbol{w}_c^{[t,n]}) \right\|^2 \right) \\
		& \quad + \sum_{c=1}^{C} G_c^2 \frac{\sigma_n^2}{p_c^2 \|\boldsymbol{h}_c\|^2}.
	\end{split}
\end{equation}

Taking the expectation of the inequality again and applying the Cauchy-Schwarz inequality, we have:
\begin{equation}
	\begin{split}
		&\mathbb{E}\left[\left\| \boldsymbol{w}^{[t+1]} - \boldsymbol{w}^{[t]} \right\|^2\right] \\
		& \leq \lambda^2 \sum_{c=1}^{C} \left( \sum_{k=1}^{K_c} G_{k,c}^2 \right) \left(G_c^2 \delta_c^2\right) \mathbb{E}\left[ \sum_{n=0}^{N_c - 1} \left\| \nabla F_c(\boldsymbol{w}_c^{[t,n]}) \right\|^2 \right] \\
		& \quad + \sum_{c=1}^{C} G_c^2 \frac{\sigma_n^2}{p_c^2 \|\boldsymbol{h}_c\|^2}.
	\end{split}
\end{equation}
From assumption \textbf{A4}, we have:
\begin{equation}
	\begin{split}
		&\mathbb{E}\left[\sum_{n=0}^{N_c - 1} \left\| \nabla F_c(\boldsymbol{w}_c^{[t,n]}) \right\|^2\right] \\
		& = \mathbb{E}\left[\left\| \nabla F_c(\boldsymbol{w}_c^{[t,0]}) \right\|^2\right]
		+ \mathbb{E}\left[\left\| \nabla F_c(\boldsymbol{w}_c^{[t,1]}) \right\|^2\right] \\
        & \quad+ \cdots + \mathbb{E}\left[\left\| \nabla F_c(\boldsymbol{w}_c^{[t,N_c-1]}) \right\|^2\right] \\
		& \quad \leq N_c \delta^2 \left\| \nabla F(\boldsymbol{w}^{[t]}) \right\|^2,
	\end{split}
\end{equation}
therefore, the following can be obtained:
\begin{equation}
	\begin{split}
		&\mathbb{E}\left[\left\| \boldsymbol{w}^{[t+1]} - \boldsymbol{w}^{[t]} \right\|^2\right] \\
		& \leq \lambda^2 \delta^2 \sum_{c=1}^{C} \left( \sum_{k=1}^{K_c} G_{k,c}^2 \right) \left(G_c^2 \delta_c^2\right) N_c \left\| \nabla F(\boldsymbol{w}^{[t]}) \right\|^2 \\
		& \quad + \sum_{c=1}^{C} G_c^2 \frac{\sigma_n^2}{p_c^2 \|\boldsymbol{h}_c\|^2}.
	\end{split}
\end{equation}

Similarly, for the second term in the upper bound of $\mathbb{E}[F(\boldsymbol{w}^{[t+1]})]$, by the Cauchy-Schwarz inequality and assumption \textbf{A4}, we have:
\begin{equation}
	\begin{aligned}
		&\mathbb{E}[(\boldsymbol{w}^{[t+1]} - \boldsymbol{w}^{[t]})^{\top} \nabla F(\boldsymbol{w}^{[t]})] \\
		&= -\lambda \mathbb{E}\left[\left(\sum_{c=1}^{C} G_c \sum_{n=0}^{N_c - 1} \sum_{k=1}^{K_c} G_{k,c} \nabla F_{k,c}(\boldsymbol{w}_c^{[t,n]}) \right)^{\top} \nabla F(\boldsymbol{w}^{[t]}) \right] \\
		& \quad + \mathbb{E}\left[\left(\sum_{c=1}^{C} G_c \frac{\sigma_n}{p_c \|\boldsymbol{h}_c\|} \right)^{\top} \nabla F(\boldsymbol{w}^{[t]})\right] \\
		&\leq -\lambda \left(\sum_{c=1}^{C} G_c \delta_c \sum_{n=0}^{N_c - 1} \sum_{k=1}^{K_c} G_{k,c} \nabla F_c(\boldsymbol{w}_c^{[t,n]}) \right)^{\top} \mathbb{E}[\nabla F(\boldsymbol{w}^{[t]})] \\
		&\leq -\lambda \delta \left(\sum_{c=1}^{C} G_c \delta_c N_c \sum_{k=1}^{K_c} G_{k,c} \nabla F(\boldsymbol{w}^{[t]}) \right)^{\top} \mathbb{E}[\nabla F(\boldsymbol{w}^{[t]})] \\
		&\leq -\lambda \delta \sum_{c=1}^{C} G_c \delta_c \sum_{k=1}^{K_c} G_{k,c} N_c \left\| \nabla F(\boldsymbol{w}^{[t]}) \right\|^2.
	\end{aligned}
\end{equation}

Therefore, the upper bound of $\mathbb{E}[F(\boldsymbol{w}^{[t+1]})]$ can be written as:
\begin{equation}
	\begin{aligned}
		&\mathbb{E}[F(\boldsymbol{w}^{[t+1]})]\\
		&\leq \mathbb{E}[F(\boldsymbol{w}^{[t]})] + \mathbb{E}[(\boldsymbol{w}^{[t+1]} - \boldsymbol{w}^{[t]})^{\top} \nabla F(\boldsymbol{w}^{[t]})] \\
        & \quad + \frac{L}{2} \mathbb{E}\left[\left\| \boldsymbol{w}^{[t+1]} - \boldsymbol{w}^{[t]} \right\|^2 \right] \\
		&\leq \mathbb{E}[F(\boldsymbol{w}^{[t]})] \\
        & \quad + \left[ \sum_{c=1}^{C} N_c \left( \frac{L}{2} \lambda^2 \delta^2 G_c^2 \delta_c^2 \sum_{k=1}^{K_c} G_{k,c}^2 - \lambda \delta G_c \delta_c \sum_{k=1}^{K_c} G_{k,c} \right) \right] \\
		&\quad \cdot \left\| \nabla F(\boldsymbol{w}^{[t]}) \right\|^2 + \frac{L}{2} \sum_{c=1}^{C} G_c^2 \frac{\sigma_n^2}{p_c^2 \|\boldsymbol{h}_c\|^2}.
	\end{aligned}
\end{equation}

From assumptions \textbf{A1} and \textbf{A3}, we have: $\|\nabla F(\boldsymbol{w}^{[t]})\|^2 \geq 2\mu [F(\boldsymbol{w}^{[t]}) - F(\boldsymbol{w}^*)]$, hence:
\begin{equation}
	\begin{aligned}
		&\mathbb{E}[F(\boldsymbol{w}^{[t+1]}) - F(\boldsymbol{w}^*)] \\
		&\leq \left[ 1 + \sum_{c=1}^{C} N_c \left( \mu L \lambda^2 \delta^2 G_c^2 \delta_c^2 \sum_{k=1}^{K_c} G_{k,c}^2 - 2 \mu \lambda \delta G_c \delta_c \sum_{k=1}^{K_c} G_{k,c} \right) \right] \\
		& \quad \cdot \mathbb{E}[F(\boldsymbol{w}^{[t]}) - F(\boldsymbol{w}^*)] + \frac{L}{2} \sum_{c=1}^{C} G_c^2 \frac{\sigma_n^2}{p_c^2 \|\boldsymbol{h}_c\|^2}.
	\end{aligned}
\end{equation}

Let
\begin{equation}
	\begin{aligned}
		A = 1 + \sum_{c=1}^{C} N_c \left( \mu L \lambda^2 \delta^2 G_c^2 \delta_c^2 \sum_{k=1}^{K_c} G_{k,c}^2 - 2 \mu \lambda \delta G_c \delta_c \sum_{k=1}^{K_c} G_{k,c} \right),
	\end{aligned}
\end{equation}
and let the FL system iterate for $T$ rounds, then we have:
\begin{equation}
\begin{aligned}
    	&\mathbb{E}[F(\boldsymbol{w}^{[t+1]}) - F(\boldsymbol{w}^*)] \\
        &\leq A^T \mathbb{E}[F(\boldsymbol{w}^{[0]}) - F(\boldsymbol{w}^*)] + \frac{1 - A^T}{1 - A} \sum_{c=1}^{C} G_c^2 \frac{\sigma_n^2}{p_c^2 \|\boldsymbol{h}_c\|^2}.
\end{aligned}
\end{equation}

Proof completed.

\section{Appendix II}

For each local iteration within a cluster, we have:
\begin{equation}
	\begin{aligned}
		&\mathbb{E}[F_c(\boldsymbol{w}_c^{[n+1]})] \\
		&\leq \mathbb{E}[F_c(\boldsymbol{w}_c^{[n]})] + \mathbb{E}[(\boldsymbol{w}_c^{[n+1]} - \boldsymbol{w}_c^{[n]})^{\top} \nabla F_c(\boldsymbol{w}_c^{[t]})] \\
        &\quad + \frac{L}{2} \mathbb{E}\left[\left\| \boldsymbol{w}_c^{[n+1]} - \boldsymbol{w}_c^{[n]} \right\|^2\right] \\
		&\leq \mathbb{E}[F_c(\boldsymbol{w}_c^{[n]})] + \mathbb{E}\left[-\lambda \delta_c \sum_{k=1}^{K_c} G_{k,c} N_c \nabla F_c(\boldsymbol{w}_c^{[t]})^{\top} \nabla F_c(\boldsymbol{w}_c^{[t]})\right] \\
		&\quad + \frac{L}{2} \lambda^2 \delta_c^2 \sum_{k=1}^{K_c} G_{k,c}^2 N_c \left\| \nabla F_c(\boldsymbol{w}_c^{[t]}) \right\|^2 \\
		&= \mathbb{E}[F_c(\boldsymbol{w}_c^{[n]})] \\
        & \quad + \left( \frac{L}{2} \lambda^2 \delta_c^2 \sum_{k=1}^{K_c} G_{k,c}^2 N_c - \lambda \delta_c \sum_{k=1}^{K_c} G_{k,c} N_c \right) \left\| \nabla F_c(\boldsymbol{w}_c^{[t]}) \right\|^2.
	\end{aligned}
\end{equation}

To ensure convergence of the intra-cluster iterations, it must hold that $\frac{L}{2} \lambda^2 \delta_c^2 \sum_{k=1}^{K_c} G_{k,c}^2 N_c - \lambda \delta_c \sum_{k=1}^{K_c} G_{k,c} N_c < 0$, that is:
\begin{equation}
	\frac{L}{2} \lambda^2 \delta_c^2 \sum_{k=1}^{K_c} G_{k,c}^2 < \lambda \delta_c \sum_{k=1}^{K_c} G_{k,c}.
\end{equation}

This inequality holds if and only if $\delta G_c < 1 < \frac{\lambda \delta_c \sum_{k=1}^{K_c} G_{k,c}}{\frac{L}{2} \lambda^2 \delta_c^2 \sum_{k=1}^{K_c} G_{k,c}^2}$, which ensures that the condition for convergence of intra-cluster updates is satisfied.

Corollary proved.

\bibliographystyle{IEEEtran}
\bibliography{references}\ 

\vfill

\end{document}